%% file: sample-sigconf.tex
\documentclass[sigconf]{acmart}
\usepackage{algorithm}
\usepackage{algorithmic}
\usepackage{amsmath}
\usepackage{hyperref}
\usepackage{makecell}
\usepackage{multirow}
\usepackage{xspace}

\usepackage{amssymb}
\usepackage{bbm}

\AtBeginDocument{%
  }

\setcopyright{acmlicensed}
\copyrightyear{2025}
\acmYear{2025}
\acmDOI{XXXXXXX.XXXXXXX}

\acmConference[Under Review]{}{August}{2024}
\begin{document}

\title{MoMQ: Mixture-of-Experts Enhances Multi-Dialect Query Generation across Relational and Non-Relational Databases}

\author{Zhisheng Lin}
\authornote{Both authors contributed equally to this research. Work done during Lin's internship at Alibaba Group.}
\affiliation{%
  \institution{Zhejiang University}
  \city{Hangzhou}
  \country{China}}
\email{linzhisheng@zju.edu.cn}

\author{Yifu Liu}
\authornotemark[1]
\affiliation{%
  \institution{Alibaba Group}
  \city{Hangzhou}
  \country{China}}
\email{zhencang.lyf@alibaba-inc.com}

\author{Zhiling Luo}
\authornote{Corresponding author.}
\affiliation{%
  \institution{Alibaba Group}
  \city{Hangzhou}
  \country{China}}
\email{godot.lzl@alibaba-inc.com}

\author{Jinyang Gao}
\affiliation{%
  \institution{Alibaba Group}
  \city{Hangzhou}
  \country{China}}
\email{jinyang.gjy@alibaba-inc.com}

\author{Yu Li}
\affiliation{%
  \institution{Alibaba Group}
  \city{Hangzhou}
  \country{China}}
\email{lojze.ly@alibaba-inc.com}

\renewcommand{\shortauthors}{Zhisheng Lin, Yifu Liu, Zhiling Luo, Jinyang Gao, Yu Li}
\newcommand{\name}{MoMQ\xspace}
% \begin{abstract}
%   A clear and well-documented \LaTeX\ document is presented as an
%   article formatted for publication by ACM in a conference proceedings
%   or journal publication. Based on the ``acmart'' document class, this
%   article presents and explains many of the common variations, as well
%   as many of the formatting elements an author may use in the
%   preparation of the documentation of their work.
% \end{abstract}
\input{section/abstract}





\maketitle
\input{section/introduction}
\input{section/related_work}
\input{section/methodology}
\input{section/experiment}
\input{section/conclusion}
\bibliographystyle{ACM-Reference-Format}
\bibliography{sample-base}

% \appendix









\end{document}

%% file: section/abstract.tex
\begin{abstract}
The improvement in translating natural language to structured query language (SQL) can be attributed to the advancements in large language models (LLMs). Open-source LLMs, tailored for specific database dialects such as MySQL, have shown great performance. However, cloud service providers are looking for a unified database manager service (e.g., Cosmos DB from Azure, Amazon Aurora from AWS, Lindorm from AlibabaCloud) that can support multiple dialects. 
This requirement has led to the concept of multi-dialect query generation, which presents challenges to LLMs. These challenges include syntactic differences among dialects and imbalanced data distribution across multiple dialects.
% To tackle these challenges, we propose MoMQ, a novel transfer learning framework for query generation across both relational and non-relational databases.
To tackle these challenges, we propose MoMQ, a novel Mixture-of-Experts-based multi-dialect query generation framework across both relational and non-relational databases.
MoMQ employs a dialect expert group for each dialect and a multi-level routing strategy to handle dialect-specific knowledge, reducing interference during query generation.
Additionally, a shared expert group is introduced to address data imbalance, facilitating the transfer of common knowledge from high-resource dialects to low-resource ones.
Furthermore, we have developed a high-quality multi-dialect query generation benchmark that covers relational and non-relational databases such as MySQL, PostgreSQL, Cypher for Neo4j, and nGQL for NebulaGraph.
Extensive experiments have shown that MoMQ performs effectively and robustly even in resource-imbalanced scenarios.
\end{abstract}

%% file: section/introduction.tex
\section{Introduction}

\begin{figure}[t]
\centering
\includegraphics[width=\linewidth]{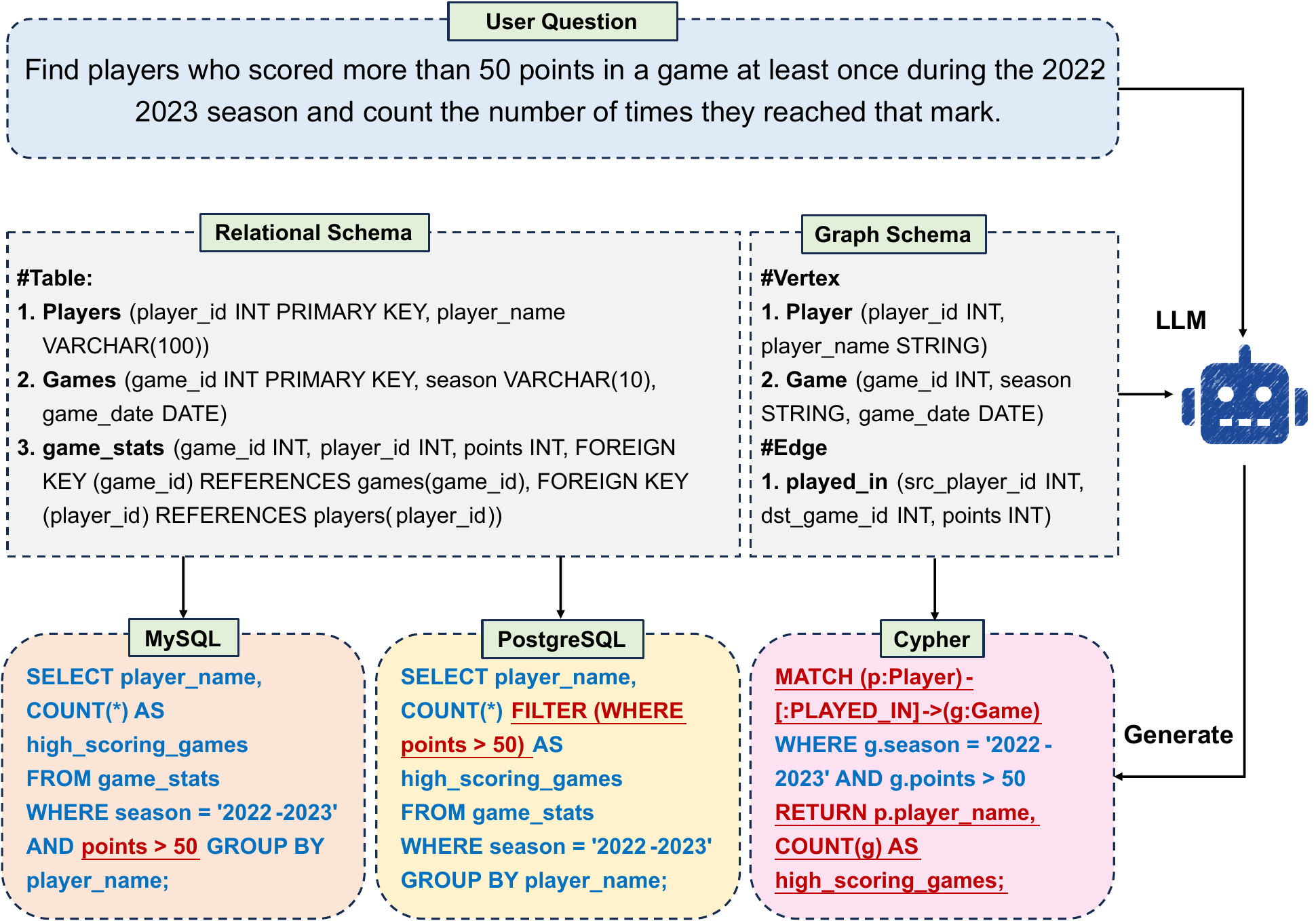}
\caption{When it comes to user queries, different database management systems may have variations in syntax. For example, PostgreSQL uses the "FILTER" keyword, while MySQL does not. Cypher uses "MATCH" for querying, while MySQL and PostgreSQL use "SELECT".}
\label{dialect_example}
\end{figure}
The ability to convert natural language into structured query language (SQL) has made it much easier to interact with relational database management systems. 
% This approach is widely used for data retrieval and analytics. 
In recent years, the use of large language models (LLMs) has significantly improved SQL generation tasks \cite{10.1145/3654930,dail-sql}. %\cite{10.1145/3654930, sun2024sqlpalmimprovedlargelanguage, pourreza2023dinsql, li2024petsqlpromptenhancedtworoundrefinement}
This shift has enhanced the quality of generated queries, moving away from encoder-decoder-based approaches \cite{10.1609/aaai.v37i11.26536, fu2023catsql, resdsql} to those driven by LLMs. Open-source LLMs \cite{10.1145/3654930, bai2023qwen, rozière2024codellamaopenfoundation} that have been fine-tuned through supervision have become the primary method due to their lower data privacy risks and cost compared to closed-source LLMs \cite{achiam2023gpt, bai2023qwen, anil2023palm}. These LLMs are typically designed to work best with a specific database dialect, like MySQL.
However, for general database management services in cloud computing, LLMs that support most dialects are needed. Therefore, SQL generation LLMs should not only cover major dialects like MySQL and PostgreSQL but also non-relational graph databases such as Neo4j \cite{neo4j} and NebulaGraph \cite{wu2022nebulagraphopensource}.
Figure \ref{dialect_example} illustrates the similarities and notable differences of the queries for the same question across different databases. 
% when a user asks a question like \textit{Find players who scored more than 50 points in a game at least once during the 2022-2023 season and count the number of times they reached that mark.} 
% The corresponding queries for the same question can have similarities and notable differences across different databases. 
The differences in relational database query languages are mainly seen in the usage of specific keywords, while the discrepancies between relational and non-relational database query languages are more distinct, reflecting differences in the underlying query logic. These differences are collectively referred to as the database dialect issue.%\cite{sqldialect}

This task is regarded as multi-dialect query generation. Past research \cite{multitask_better, multitask_better2, multitask_better3} has shown that training models on multiple tasks can help them integrate knowledge from different sources, leading to improved performance compared to training on a single task. However, directly fine-tuning dense LLMs on multi-dialect data encounters several challenges:

1. There are syntax differences between relational database dialects, such as the use of specific keywords in MySQL and PostgreSQL. Additionally, the disparities between relational and non-relational databases are more pronounced and are reflected in the query logic, such as "SELECT" statements in MySQL and "MATCH" statements in Cypher.
Neglecting these differences between dialects may interfere with the generation of accurate queries.

2. The cost of annotating natural language to database query language is significant, and data for most dialects is limited. Importantly, the similarities in natural language questions and database schemas across dialects represent transferable knowledge that is not fully utilized. It would be beneficial to transfer this common knowledge from high-resource dialects to low-resource ones to address data imbalance.

To tackle these challenges, we propose \name, a \underline{M}ixture-\underline{o}f-Experts (MoE)-based \underline{m}ulti-dialect \underline{q}uery generation framework. This framework unifies the generation of query languages for both relational and non-relational databases. 
Building MoE structures from scratch through pre-training LLMs is expensive, so we construct the MoE based on a dense model.
In contrast to MoE Upcycling \cite{komatsuzakisparse}, we use multiple Low-Rank Adaptation (LoRA) \cite{hu2021lora} modules to develop a detailed MoE structure, as in \cite{wu2024mixtureloraexperts, li2024mixloraenhancinglargelanguage, feng2024mixtureoflorasefficientmultitasktuning}. Meanwhile, we freeze the original model to prevent the loss of pre-trained knowledge.
We create specialized dialect expert groups for each dialect to isolate dialect-specific knowledge and reduce interference during generation. To address the imbalance in multi-dialect data, we introduce a shared expert group visible to all dialects, enhancing the transfer of common knowledge from high-resource dialects to low-resource dialects.
We further propose a novel multi-level routing strategy consisting of a dialect router and an expert router. 
The dialect router, on the one hand, routes dialect-specific tokens to their respective dialect groups. 
On the other hand, with the assistance of the Dialect Router Loss, it distributes other tokens across all groups to facilitate token-level knowledge transfer. 
% On the other hand, it evenly distributes other tokens across all groups.
% The dialect router will effectively route tokens of various dialects to the appropriate dialect expert group. 
% To enable similar tokens to be processed by diverse dialect expert groups, we introduce the Dialect Router Loss, which facilitates token-level knowledge transfer. 
The expert router is responsible for activating the top-k experts within a dialect expert group for input tokens. 
To prevent load imbalance, which is known as routing collapse \cite{Sparsely-Gated}, we also utilize an Expert Balance Loss.

We conduct a thorough evaluation of the query language generation capability across various databases. We construct a high-quality benchmark that covers both relational and non-relational graph databases, such as MySQL, PostgreSQL, Cypher for Neo4j, and nGQL for NebulaGraph.
The evaluation of our approach is carried out across various scenarios, including full data and data-imbalanced settings. 
The results show that in the full data setting, our approach improved execution accuracy by an average of 3-5 percent compared to the baselines across all database dialects. 
In data-imbalanced settings, the average improvement in execution accuracy is 4-6 percent for most dialects. 
These findings confirm the effectiveness of our approach in handling dialectal interference and data imbalance across diverse database dialects.

We further conduct a series of analytical experiments to assess the impact of the capacity and number of experts on the performance of our approach, demonstrating its robustness. 
Additionally, we analyze the distribution of expert weights to validate our approach's ability to capture multi-dialect knowledge and transfer generalized dialect knowledge.
Both the code and multi-dialect benchmark will be openly released to support further research in this area. Overall, we summarize our contributions as follows:
\begin{itemize}
	\item We introduce \name, a MoE-based multi-dialect transfer learning framework for query generation across both relational and non-relational databases. This framework significantly enhances the multi-dialect generation capabilities of open-source large models.
	\item We propose a dialect-specific expert group structure with a multi-level routing mechanism that explicitly models the specialization capabilities of different dialect experts and adaptively achieves the transfer of shared knowledge between dialects.	
	\item To better evaluate the model’s performance in multi-dialect query generation, we construct a high-quality multi-dialect benchmark. This benchmark covers both relational and non-relational graph database dialects, including MySQL, PostgreSQL, Cypher for Neo4j, and nGQL for NebulaGraph.
	\item The experimental results show that our approach, \name, significantly improves multi-dialect query generation, even in data-imbalanced settings. This highlights the effectiveness of our method. Further analytical experiments support the effectiveness of our approach by examining the performance differences of \name under various settings.
\end{itemize}

%% file: section/related_work.tex
\section{Related Work}
% \subsection{Mixture-of-Experts}
% \subsection{Text-to-SQL}
% \subsection{Multi-task Learning}
% \subsection{Supervised Fine-Tuning-Based Text-to-SQL}
\begin{figure*}[t]
\centering
\includegraphics[width=0.9\linewidth]{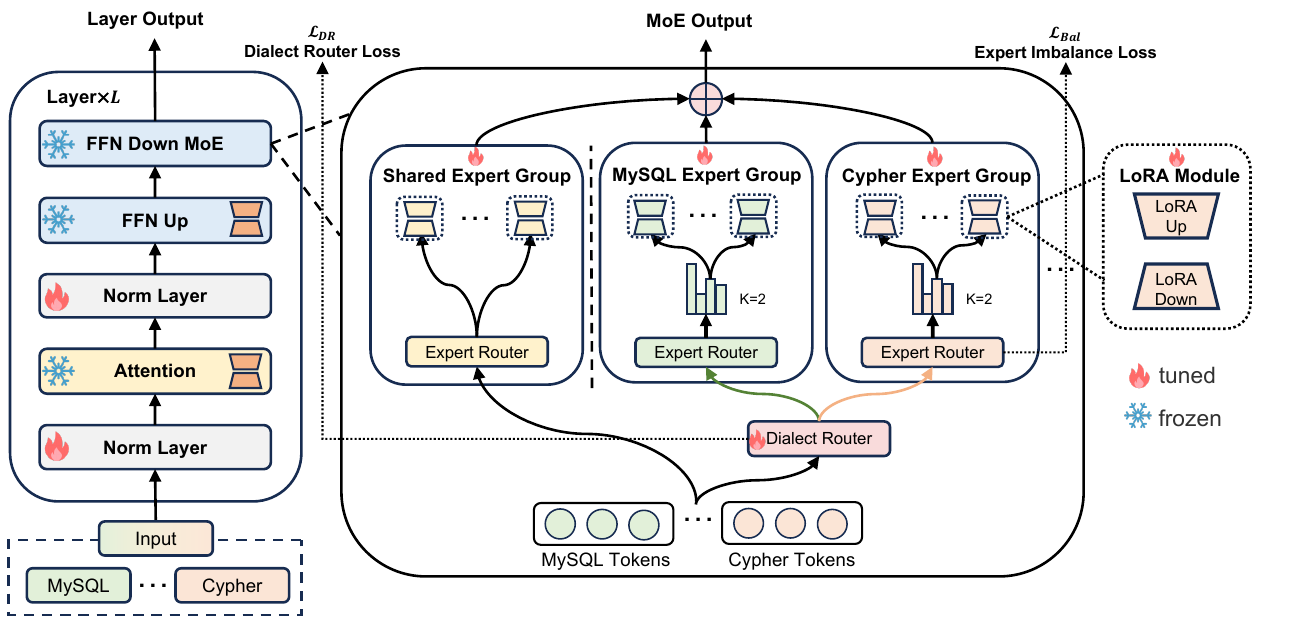}
\caption{The overall structure of \name. The original feed-forward network (FFN) is transformed into a MoE structure, which consists of Shared Expert Group, Dialect Expert Group, and Multi-Level Strategy. The pre-trained weights are frozen and LoRA modules inserted into Attention and FFN are fine-tuned for rapid adaptation to multi-dialect query generation. The normalization layer is unfrozen due to its observed improvement. The Dialect Router Loss and Expert Balance Loss are added to the training objectives to adjust multi-dialect routing and mitigate routing collapse respectively.}
\label{architecture}
\end{figure*}

% In this work, we propose a multi-dialect query generation framework based on the MoE model, related to the query generation and the Mixture-of-Experts research.
% The main work of query generation has focused on the field of text-to-SQL generation.
\name is a multi-dialect query generation framework that is highly related with the tuning-based text-to-SQL task and Mixture-of-Experts.
\subsection{Tuning-Based Text-to-SQL} Before the era of large-scale models, the text-to-SQL task typically employs an encoder-decoder-based architecture. 
Research in this domain primarily focuses on enhancing the encoder's ability to understand the question and schema \cite{10.1609/aaai.v37i11.26536, fu2023catsql, resdsql}.
With the advent of large language models (LLMs), the text-to-SQL task has gradually shifted towards LLM-based approaches\cite{achiam2023gpt, bai2023qwen, anil2023palm, 10.1145/3654930, li2024petsqlpromptenhancedtworoundrefinement, pourreza2023dinsql}. Supervised fine-tuning of open-source large models such as LLama \cite{touvron2023llamaopenefficientfoundation}, StartCoder \cite{li2023starcodersourceyou}, and Qwen \cite{bai2023qwen} has significantly improved both natural language understanding and SQL query generation capabilities.

\subsection{Mixture-of-Experts}
In recent years, Mixture-of-Experts (MoE) has emerged as an effective structure for reducing inference computational costs and enhancing multi-task learning capabilities in scenarios where model parameters are continuously scaled up \cite{dai2024deepseekmoeultimateexpertspecialization, jiang2024mixtralexperts, zhao2024sparse, xue2024openmoeearlyeffortopen, 9835248, fedus2022switchtransformersscalingtrillion, lepikhin2020gshardscalinggiantmodels, pmlr-v162-du22c}.
A gate unit is then utilized for expert activation.
However, training MoE models from scratch or converting dense models into MoE through upcycling requires substantial pre-training costs \cite{komatsuzakisparse, lin2024moe, xue2024openmoeearlyeffortopen}.
Recent studies have proposed the construction of MoE based on the Low-Rank Adaptation (LoRA) module \cite{wu2024mixtureloraexperts, li2024mixloraenhancinglargelanguage, feng2024mixtureoflorasefficientmultitasktuning}.
This approach not only mitigates catastrophic forgetting of pre-trained knowledge in dense models but also enhances the multi-task learning capability.

%% file: section/methodology.tex
\section{Methodology}
\subsection{Task Definition}
Given $\mathcal{N}$ natural language questions $\mathcal{Q} = \{q_1,q_2\cdots,q_\mathcal{N}\}$, a set of target query language dialects $\mathcal{L} = \{l_1,l_2\cdots,l_\mathcal{M}\}$ and a database schema $\mathcal{D}=\{\mathcal{C}, \mathcal{T}\}$, where $\mathcal{C}$ and $\mathcal{T}$ represent columns and tables.
Formally, the goal is to learn a mapping function as:
\begin{equation}
\mathcal{F}: (\mathcal{Q}, \mathcal{L}, \mathcal{D}) \rightarrow \mathcal{S},
\end{equation}
where for each input question $q_i \in \mathcal{Q}$, schema $\mathcal{D}$ and dialect $l_j \in \mathcal{L}$, the function $\mathcal{F}$ generates a valid database query $\mathcal{S}_{i,j}$.
\subsection{Architecture Overview}
% To fully isolate dialect-specific knowledge from multi-dialect data and enhances dialect-shared knowledge transfer, we transform dense model to token-level MoE structure by replacing FFN Down Layer with MoE layer.
The overall architecture of \name is illustrated in Figure \ref{architecture}.
% Each expert group contains several fine-grained LoRA modules. 
% \name leverages LoRA modules as fine-grained experts to construct expert groups.
\name constructs MoE structure by leveraging LoRA modules as fine-grained experts.
% Specifically，
% We introduce dialect expert groups for each dialect and a shared expert group visible to all dialects.
Dialect expert groups for each dialect and a shared expert group visible to all dialects are further introduced.
% In the multi-level routing strategy, the dialect routers and expert routers collaboratively ensure the correct routing from the dialect expert groups to within the group.
The multi-level routing strategy, composed of a dialect router and an expert router, ensures the correct routing of tokens between different expert groups and within each group.
When the tokens of different dialects are input, they are initially directed through the dialect router to the appropriate dialect expert groups. 
Within these groups, the tokens are further routed by the expert router to the final activated experts. 
Notably, all tokens are fully processed by the shared expert group, which is beneficial for the fusion and transfer of multi-dialect knowledge. 
Besides, to better assist the dialect router in effectively routing tokens of various dialects to the appropriate expert group, we introduce a novel Dialect Router Loss. 
% This loss is fundamentally designed to perform a token-level multi-class classification task. 
% It takes the router outputs and the dialect category labels of the tokens and outputs the expert group categories. Additionally, we incorporate a practical Expert Balance Loss to mitigate the risk of routing collapse.

% In the rest of this section, we introduce the construction of MoE, the design of expert groups (including dialect expert groups and a shared expert group),
% and the dialect router used for different dialect groups.
% Finally, we explain the training objectives of \name.

% In the following sections, we present the construction of the MoE structure, along with the design of the expert groups, including dialect-specific expert groups and a shared expert group. Additionally, we describe the dialect router that manages the routing between different dialect groups.

In the following sections, we first present the construction of the MoE structure, along with the design of the expert groups and dialect router.
Then, we outline the training objectives of \name.

\subsection{MoE Construction}
% \subsection{LoRA Module.}
% Low-Rank Adaptation (LoRA) has been widely used to adapt large pre-trained models to specific tasks or datasets efficiently \cite{hu2021lora}. The LoRA approach introduces matrices, $\mathbf{B}\in \mathbb{R}^{m\times r}$ and $\mathbf{A}\in {\mathbb{R}}^{r\times n}$, where $r \ll \min(m, n)$, and defines the adapted weight matrix \(W'\) as:
% \begin{equation}
% \mathbf{W^\prime} = \mathbf{W_0} + \mathbf{B}\mathbf{A}, 
% \end{equation}
% where $\mathbf{W_0} \in \mathbb{R}^{m\times n}$ is the original pre-trained weight matrix that remains fixed during the adaptation.
% \subsection{MoE Construction.}
MoE structures are constructed in a variety of ways, such as pre-training from scratch or replicating multiple FFNs followed by a step of continual pre-training, all of which require additional pre-training to inject knowledge into the MoE \cite{komatsuzakisparse, lin2024moe, xue2024openmoeearlyeffortopen}.  
Inspired by recent works \cite{wu2024mixtureloraexperts, li2024mixloraenhancinglargelanguage, feng2024mixtureoflorasefficientmultitasktuning}.
We use a simple but efficient way to construct MoE, freezing the original LLMs and inserting multiple Low-Rank Adaptation (LoRA) \cite{hu2021lora} modules into FFN as fine-grained experts.
Specifically, a transformer FFN consists of two stacked layers, an up-projection layer and a down-projection layer. We replace the down-projection layer with multiple LoRA modules and deploy vanilla LoRA modules in both the up-projection layer and the attention layer to facilitate the model's rapid adaptation to multi-dialect query generation.
% Low-Rank Adaptation (LoRA) has been widely used to adapt large pre-trained models to specific tasks or datasets efficiently \cite{hu2021lora}.
The LoRA module consists of two matrices, $\mathbf{B}\in \mathbb{R}^{m\times r}$ and $\mathbf{A}\in {\mathbb{R}}^{r\times n}$, where $r \ll \min(m, n)$, and defines the adapted weight matrix $\mathbf{W^\prime}$ as:
\begin{equation}
\mathbf{W^\prime} = \mathbf{W_0} + \mathbf{B}\mathbf{A}, 
\end{equation}
where $\mathbf{W_0} \in \mathbb{R}^{m\times n}$ is the original pre-trained weight matrix that remains fixed during the adaptation.
All LoRA modules work parallel to the original layer to fully utilize the pre-trained knowledge.

\subsection{Dialect Expert Group}
In multi-dialect generation, there are non-trivial syntax differences between two database dialects.
For example, in MySQL, the "DATE\_ADD" function can be used to add a specified time interval to a date or datetime value. While PostgreSQL uses the "INTERVAL" keyword along with the "+" operator to achieve a similar result.
These differences will interfere with learning dialect-specific knowledge.
% In dense models, multiple dialects share the same set of model parameters, which could lead to interference during generation.

To address this issue, we design multiple dialect expert groups to isolate dialect-specific knowledge, thereby mitigating interference and improving the quality of generated queries.
The expert group consists of multiple LoRA experts and a top-k expert router. 
% We employ two types of expert groups, dialect expert group and shared expert group.
Each database dialect has a separate expert group to learn knowledge of the corresponding query syntax.
Concretely, given multiple dialect expert groups $\mathcal{G}=\{G_1, G_2, \cdots, G_M\}$ in a specific Transformer layer, the output of the $i$-th expert group is calculated as:
\begin{equation}
	\mathbf{H}_e^i = \sum_{k=1}^{N}g_{k}^{i}\cdot\mathbf{E}_{k}^{i},
\end{equation}
\begin{equation}
g_{k}^{i}=
\begin{cases}
    s_{k}^{i}, & s_{k}^{i}\in\text{TopK}(\{s_{k}^{i}|1 \leq j \leq N\}, K) \\
    0, & \text{otherwise}
\end{cases},
\end{equation}
\begin{equation}
	s_{k}^{i} = {\rm softmax}(\mathbf{W}_e^i \cdot \mathbf{H})_k,
	\label{eqn:task_corr}
\end{equation}
where $\mathbf{E}_{k}^{i}$ is the $k$-th LoRA expert in the $i$-th expert group, and $N$ is the total number of experts in the group,
$g_{k}^{i}$ denotes the $k$-th gate value for the expert, $s_{k}^{i}$ denotes the token-to-expert affinity, $\text{TopK}(\cdot, K)$ denotes the set comprising $K$ highest affinity scores among those calculated for the tokens in all $N$ experts, $\mathbf{W}_e^i \in \mathbb{R}^{d \times K}$ is a trainable matrix of the expert router, and $\mathbf{H}$ is the hidden states of all input tokens input. Note that $g_{k}^{i}$ is sparse, indicating that only $K$ out of $N$ gate values are nonzero. This sparsity property ensures computational efficiency within an expert group, i.e., each token will be assigned to and computed in only $K$ experts.

The routing strategy within the expert group faces the problem of load imbalance \cite{Sparsely-Gated}. This can lead to a situation known as routing collapse, where the expert router continually selects a limited set of experts, thereby inhibiting adequate training for the others.
In order to mitigate the risk of routing collapse, we employ an expert-level balance loss, which is computed as follows:

\begin{equation}
\mathcal{L}_{Bal} = N\cdot\sum_{i=1}^{N} f_i \cdot P_i,
\end{equation}
\begin{equation}
% f_i = \frac{N'}{K'T} \sum_{t=1}^T \mathbbm{1}(Token~t~selects~Expert~i),
f_i = \frac{1}{KT} \sum_{t=1}^T \mathbbm{1}(Token~t~selects~Expert~i),
\end{equation}
\begin{equation}
P_i = \frac{1}{T} \sum_{t=1}^T s_{i,t},
\end{equation}
where $T$ is the number of processed tokens, $\mathbbm{1}$ is the indicator function, $f_i$ is the fraction of tokens dispatched to expert $i$ and $P_i$ is the fraction of the router probability allocated for expert $i$.

\begin{table*}[ht]
    \centering
    \scalebox{1}{
    \begin{tabular}{cc|ccccc|ccccc}
        \Xhline{1pt}
        \multirow{2}{*}{\textbf{Backbone}} & \multirow{2}{*}{\textbf{Method}} & \multicolumn{5}{c}{\textbf{Execution Accuracy}} & \multicolumn{5}{c}{\textbf{Executable}} \\
        \cline{3-12}
         & & \textbf{MySQL} & \textbf{PG} & \textbf{Cypher} & \textbf{nGQL} & \textbf{Avg.} & \textbf{MySQL} & \textbf{PG} & \textbf{Cypher} & \textbf{nGQL} & \textbf{Avg.} \\
        \Xhline{1pt}
        \multirow{4}{*}{Qwen2-1.5B}&Full Fine-Tuning& 43.30 & 27.00 & 23.61 & 22.92 & 29.21 & 64.21 & 66.67 & 91.55 & 65.97 & 72.10 \\
        &Full Fine-Tuning$^*$& 27.80 & 24.30 & 21.30 & 9.61 & 20.75 & 45.63 & 67.37 & 87.15 & 62.38 & 65.64 \\
        &LoRA& 49.20 & 32.16 & 25.12 & 26.97 & 33.36 & 67.28 & 69.84 & \textbf{92.94} & 71.53 & 75.40 \\
        &\textbf{\name(Ours)} & \textbf{52.15} & \textbf{32.75} & \textbf{27.55} & \textbf{30.21} & \textbf{35.66} & \textbf{70.23} & \textbf{71.01} & 91.78 & \textbf{82.18} & \textbf{78.80} \\
        \Xhline{1pt}
        \multirow{4}{*}{Qwen2-7B}&Full Fine-Tuning& 63.71 & 46.24 & 39.12 & 36.57 & 46.41 & 83.52 & 82.75 & \textbf{96.53} & 85.07 & 86.97 \\
        &Full Fine-Tuning$^*$& 58.92 & 44.37 & \textbf{41.44} & 38.66 & 45.84 & 78.72 & \textbf{85.09} & 95.95 & 85.65 & 86.35 \\
        &LoRA& 63.35 & 44.95 & 38.31 & 27.31 & 43.48 & 84.87 & 83.80 & 95.83 & 82.52 & 86.76 \\
        &\textbf{\name(Ours)} & \textbf{66.30} & \textbf{48.12} & 40.97 & \textbf{41.20} & \textbf{49.15} & \textbf{87.21} & 84.51 & 95.83 & \textbf{87.38} & \textbf{88.73} \\
        \Xhline{1pt}
        \multirow{4}{*}{Qwen1.5-14B}&Full Fine-Tuning& 46.68 & 45.95 & 36.46 & 28.65 & 39.43 & 67.16 & 80.99 & 93.23 & 80.90 & 80.57 \\
        &Full Fine-Tuning$^*$& 34.93 & 44.01 & 37.85 & 28.24 & 36.26 & 53.51 & \textbf{82.51} & 93.52 & 68.87 & 74.60 \\
        &LoRA& 50.55 & 45.89 & 36.11 & 28.59 & 40.29 & 70.85 & 81.93 & 91.67 & 79.17 & 80.90 \\
        &\textbf{\name(Ours)} & \textbf{61.75} & \textbf{47.65} & \textbf{39.47} & \textbf{34.26} & \textbf{45.78} & \textbf{81.55} & 81.69 & \textbf{94.91} & \textbf{82.18} & \textbf{85.08} \\
        \Xhline{1pt}
    \end{tabular}
    }
    \caption{Results in the full data setting. PG refers to PostgreSQL and * denotes single-dialect full fine-tuning.}
    \label{full-data}
\end{table*}

\subsection{Shared Expert Group}
% With a conventional routing strategy, tokens are assigned to a specific dialect expert group independently. 
% There is no inherent information communication between different dialects, thus may yield a negative impact on common knowledge transfer from high-resource dialects to low-resource dialects, especially in data-imbalanced settings.
There is a lack of inherent information communication between different dialects by using only a separate expert group for each dialect.
When encountering data imbalance, it can negatively impact the common knowledge transfer from high-resource dialects to low-resource dialects.
% This can have a negative impact on the transfer of common knowledge from high-resource dialects to low-resource dialects when data imbalance is encountered.
Meanwhile, multiple experts may converge in acquiring shared knowledge in their respective parameters, thereby resulting in redundancy in expert parameters.
If there are shared experts dedicated to capturing and consolidating common knowledge across varying dialects, knowledge transfer will be more efficient and parameter redundancy will be alleviated.

Toward this objective, we further add a shared expert group to integrate information across multiple dialects at the sentence level.
Regardless of the router module, all tokens in a sentence will be deterministically assigned to experts in the shared expert group. Formally, the MoE output in the complete \name architecture is formulated as follows:

\begin{equation}
	\mathbf{H}_o = \sum_{i=1}^{M}\mathbf{H}_{e}^{i} + \sum_{k=1}^{N_s}\mathbf{E}_{k}^s,
\end{equation}
where $M$ is the number of dialect expert groups, $N_s$ is the number of shared experts and $\mathbf{E}_{k}^s$ is the output of $k$-th shared expert.

\subsection{Dialect Router}
% With a conventional routing strategy, tokens are assigned to a specific dialect expert group independently. 
% After dividing the multi-dialect expert group, how to make the tokens of different dialects routed to the correct dialect expert group is an important issue that needs to be solved. 
After the construction of multiple dialect expert groups, how to route the tokens of different dialects to the appropriate group remains to be addressed.
Intuitively, the dialect router is able to make correct routing under the guidance of sentence-level dialect hard labels, thus forming a complete isolation between different dialect expert groups.
However, there may exist certain similarities between different dialects, e.g., "LIMIT" and "ORDER BY" in relational and non-relational database dialects are both valid tokens. Moreover, different dialects exhibit a high degree of similarity in the understanding of natural language questions and database schemas.
If these similar tokens have the opportunity to enter multiple dialect expert groups, especially from high-resource dialects to low-resource ones, may further facilitate token-level common knowledge transfer.

To this end, we have designed a novel dialect router trained with a Dialect Router Loss (DRL) incorporating dialect smoothing.
Dialect smoothing is employed to further reduce dialect isolation by replacing hard dialect labels with a smooth distribution.
This distribution assigns a lower value to the true dialect while allocating a portion of the value mass to other dialects.
Under the constraint of DRL, tokens excluding dialect-specific ones have a higher probability of being routed to various dialect experts group upon the output weight of the dialect router, thus facilitating more comprehensive dialect information exchange and knowledge transfer. 
We add the DRL to the training objective and it is formally defined as:
\begin{equation}
\tilde{y}_t=y_t(1-\epsilon)+\frac{\epsilon}{M},
\end{equation}
\begin{equation}
	\mathcal{L}_{DR} = -\frac{1}{L T}\sum_{l=1}^{L}\sum_{t=1}^{T} r_t^l\log(\tilde{y}_t),
\end{equation}
where $y_t$ represents one-hot vector label among $M$ dialect groups, $\epsilon \in [0,1]$ is the smoothing factor, $\tilde{y}_t$ is the label after smooth, $r_t^l$ denotes the output logits of dialect router in the $l$-th Transformer layer for the $t$-th token, $L$ is the total number of layers, $T$ it the total number of input tokens, and $y_t$ is the dialect class label.

\subsection{The Training Objectives}
% \subsection{Expert Balance Loss}
% \subsection{Fine-Tuning on Query Generation Task}
Finally, we formulate the multi-dialect query generation task as a text-to-text problem. The training objective is to minimize the negative log-likelihood of output $\mathbf{y}$ conditioned on the input question $\mathbf{x}$ and the task prompt $\mathbf{P}$. The fine-tuning loss on the task is defined as:
\begin{equation}
	\mathcal{L}_{FT} = -\sum_{j}\mathcal{P}(y_j|\mathbf{y}_{<j};\mathbf{x}, \mathbf{P}) + \alpha\mathcal{L}_{DR} + \lambda\mathcal{L}_{Bal},
\end{equation}
where $\alpha$ and $\lambda$ are hyper-parameters that are used to adjust the impact of auxiliary losses.

%% file: section/experiment.tex
\section{Experiments}
\subsection{Datasets}

There is currently no unified benchmark for evaluating the performance of LLMs in multi-dialect query generation. For the text-to-SQL task, there are several widely used benchmarks, such as Spider \cite{yu2018spider} and BIRD \cite{li2024can} in English, as well as Chase \cite{guo-etal-2021-chase} in Chinese.
All of these benchmarks are constructed based on SQLite.
Additionally, there are multilingual benchmarks like MultiSpider \cite{dou2023multispider}, designed to assess the multilingual comprehension capability of the model. Consequently, we developed a training and evaluation dataset for converting natural language to query language of both relational and non-relational databases in English and Chinese.

As shown in Table \ref{tab:dialect_data}, our benchmark contains a total of four dialects.
\begin{table}[ht]
    \centering
    \begin{tabular}{cccccc}
        \Xhline{1pt}
        & \textbf{MySQL} & \textbf{{PG}} & \textbf{Cypher} & \textbf{nGQL}\\
        \hline
        \textbf{Train} & 3,000 & 3,000 & 3,000 & 3,000 \\
        \textbf{Test} & 280 & 304 & 288 & 288 \\
        \Xhline{1pt}
    \end{tabular}
    \caption{The statistics of data for each dialect, including MySQL, PostgreSQL (PG), Cypher for Neo4j, and nGQL for NebulaGraph.}
    \label{tab:dialect_data}
\end{table}
For MySQL, we each sampled 1,500 examples from the training sets of Spider and Chase for training. Additionally, we selected the "superhero" and "student\_club" databases from the dev set of BIRD as the test set. The syntax of all samples was accurately converted from SQLite to MySQL.

Regarding PG, we obtained 3,000 samples from BIRD, transforming the original SQLite syntax into PostgreSQL syntax for the training set. As for the test set, we directly used SQL-Eval\footnote{https://github.com/defog-ai/sql-eval}, an evaluation set released by Defog. It's based on the schema from Spider but with a new set of hand-selected questions and queries grouped by query category. 

For Cypher, the samples were acquired from the open-source text-to-cypher dataset\footnote{https://github.com/neo4j-labs/text2cypher}, which includes over 16 different graph schemas, along with graph information for evaluation. We sampled "fincen" and "movies" databases as the test set and used the remainder for training.

In the case of nGQL, we utilized a dataset published by \citet{zhou2024r3nl2gqlmodelcoordinationknowledge}. It involves matching data from different Knowledge Graphs to the NebulaGraph format, as well as generating training and testing data. 
All datasets for the aforementioned dialects underwent manual inspection and filtering to ensure the executability and correctness of each query.

\begin{table*}[ht]
    \centering
    \scalebox{1}{
    \begin{tabular}{cc|ccccc|ccccc}
        \Xhline{1pt}
        \multirow{2}{*}{\textbf{Backbone}} & \multirow{2}{*}{\textbf{Method}} & \multicolumn{5}{c}{\textbf{Execution Accuracy}} & \multicolumn{5}{c}{\textbf{Executable}} \\
        \cline{3-12}
         && \textbf{MySQL} & \textbf{PG} & \textbf{Cypher} & \textbf{nGQL} & \textbf{Avg.} & \textbf{MySQL} & \textbf{PG} & \textbf{Cypher} & \textbf{nGQL} & \textbf{Avg.} \\
        \Xhline{1pt}
        \multirow{3}{*}{Qwen2-1.5B}&Full Fine-Tuning & 30.87 & 23.76 & 15.74 & 3.13 & 18.37 & 48.46 & 56.50 & 80.79 & 59.38 & 61.28 \\
        &LoRA & 35.55 & \textbf{27.30} & 17.82 & 4.17 & 21.21 & 53.01 & \textbf{60.28} & 85.42 & 48.73 & 61.86 \\
        &\textbf{\name (Ours)} & \textbf{39.73} & 24.35 & \textbf{19.21} & \textbf{6.25} & \textbf{22.39} & \textbf{60.27} & 60.05 & \textbf{86.46} & \textbf{59.72} & \textbf{66.62} \\
        \hline
        \multirow{3}{*}{Qwen2-7B}&Full Fine-Tuning & 60.27 & 45.39 & 27.66 & \textbf{11.46} & 36.20 & 78.84 & \textbf{82.03} & 92.36 & \textbf{50.35} & \textbf{75.90} \\
        &LoRA & 61.38 & 43.62 & 27.32 & 3.94 & 34.06 & \textbf{81.30} & 79.43 & 91.90 & 33.91 & 71.64 \\
        &\textbf{\name (Ours)} & \textbf{63.22} & \textbf{46.93} & \textbf{28.59} & 9.95 & \textbf{37.17} & \textbf{81.30} & 81.32 & \textbf{92.82} & 45.37 & 75.21 \\
        \hline
        \multirow{3}{*}{Qwen1.5-14B}&Full Fine-Tuning & 39.98 & 41.25 & 26.04 & 5.79 & 28.26 & 57.69 & 71.75 & \textbf{90.28} & 62.04 & 70.44 \\
        &LoRA & 43.05 & 41.02 & 23.03 & 1.74 & 27.21 & 60.39 & 70.45 & 82.41 & 31.48 & 61.18 \\
        &\textbf{\name (Ours)} & \textbf{53.38} & \textbf{45.39} & \textbf{26.50} & \textbf{13.54} & \textbf{34.70} & \textbf{74.91} & \textbf{75.77} & 87.50 & \textbf{64.58} & \textbf{75.69} \\
        \Xhline{1pt}
    \end{tabular}
    }
    \caption{Results in the MySQL high-resource setting. All available data from MySQL is utilized, while 128 examples are sampled from each of the other dialects.}
     \label{mysql-rich}
\end{table*}

\begin{table*}[ht]
    \centering
    \scalebox{1}{
    \begin{tabular}{cc|ccccc|ccccc}
        \Xhline{1pt}
        \multirow{2}{*}{\textbf{Backbone}}&\multirow{2}{*}{\textbf{Method}} & \multicolumn{5}{c}{\textbf{Execution Accuracy}} & \multicolumn{5}{c}{\textbf{Executable}} \\
        \cline{3-12}
         && \textbf{MySQL} & \textbf{PG} & \textbf{Cypher} & \textbf{nGQL} & \textbf{Avg.} & \textbf{MySQL} & \textbf{PG} & \textbf{Cypher} & \textbf{nGQL} & \textbf{Avg.} \\
        \Xhline{1pt}
        \multirow{3}{*}{Qwen2-1.5B}&Full Fine-Tuning & 28.41 & \textbf{27.42} & 21.76 & 1.62 & 19.80 & 44.40 & 54.61 & 87.73 & 49.65 & 59.10 \\
        &LoRA & 31.98 & 25.77 & 24.77 & 3.47 & 21.50 & 48.71 & 59.46 & 91.90 & 49.19 & 62.31 \\
        &\textbf{\name (Ours)} & \textbf{33.09} & 22.81 & \textbf{26.04} & \textbf{5.79} & \textbf{21.93} & \textbf{49.20} & \textbf{64.18} & \textbf{92.59} & \textbf{55.90} & \textbf{65.47} \\
        \hline
        \multirow{3}{*}{Qwen2-7B}&Full Fine-Tuning & 61.13 & \textbf{47.99} & 43.17 & 5.67 & 39.49 & 78.48 & 81.68 & \textbf{96.18} & 47.57 & 75.98 \\
        &LoRA & \textbf{62.12} & 50.35 & 38.66 & 4.17 & 38.82 & \textbf{80.32} & \textbf{83.69} & 94.56 & 36.46 & 73.76 \\
        &\textbf{\name (Ours)} & \textbf{62.12} & 48.46 & \textbf{44.68} & \textbf{10.53} & \textbf{41.45} & 78.84 & 82.39 & 95.72 & \textbf{56.02} & \textbf{78.24} \\
        \hline
        \multirow{3}{*}{Qwen1.5-14B}&Full Fine-Tuning & 36.65 & \textbf{47.99} & 38.08 & 3.70 & 31.61 & 54.12 & 78.49 & 94.33 & 60.30 & 71.81 \\
        &LoRA & \textbf{46.99} & 43.50 & 31.02 & 2.43 & 30.98 & \textbf{63.10} & 72.22 & 90.39 & 50.00 & 68.93 \\
        &\textbf{\name (Ours)} & 43.42 & 46.45 & \textbf{40.05} & \textbf{10.19} & \textbf{35.03} & 57.32 & \textbf{79.67} & \textbf{94.79} & \textbf{65.86} & \textbf{74.41} \\
        \Xhline{1pt}
    \end{tabular}
    }
    \caption{Results in the Cypher high-resource setting. All available data from Cypher is utilized, while 128 examples are sampled from each of the other dialects.}
    \label{cypher-rich}
\end{table*}

\subsection{Experimental Setup}
\subsubsection{Evaluation Metric.}
We consider two prevalent evaluation metrics: execution accuracy (EX) and executable (EXEC). The EX metric evaluates whether the predicted and ground-truth queries yield the same execution results on the database. The EXEC metric evaluates whether the generated query can be executed correctly in the corresponding database without syntax errors.

\subsubsection{Models and Baselines.}
\name is a generic multi-task framework, especially for multi-dialect query generation, that can be used on a variety of open-source LLMs.
We select three model sizes of current SOTA models, namely Qwen2-1.5B, Qwen2-7B, and Qwen1.5-14B from HuggingFace\footnote{https://huggingface.co/}, as backbones respectively to validate \name's multi-dialect query generation capabilities and robustness.
Note that Qwen2 does not currently have open-source 14B weights, so we select Qwen1.5-14B as an alternative. All models use the Instruct or Chat version instead of the Base version to obtain better performance.

We compare \name with the following methods: (1) Single-dialect full fine-tuning, which fine-tunes all parameters of the model in each dialect data; (2) Multi-dialect full fine-tuning, which fine-tunes in a mixed dataset of multiple dialects; (3) Vanilla LoRA, which freezes the pre-trained model and replaces all linear layers with LoRA modules in a mixed dataset of multiple dialects.

\subsubsection{Implementation Details.}
Our experiments are conducted using PyTorch 2.3.1 on a computer running the Ubuntu 20 operating system, equipped with 8 NVIDIA A100 80GB GPUs.
We establish a shared expert group with 2 LoRA experts and four dialect-specific expert groups corresponding to MySQL, PostgreSQL, Cypher, and nGQL.
Each dialect expert group comprises 8 LoRA experts, with each input token activating the top-2 experts.
For models with different parameter sizes, we employ varying expert dimensions: 64 dimensions for the 1.5B model, 128 dimensions for the 7B model, and 256 dimensions for the 14B model.
To optimize the objectives, we use the AdamW optimizer with parameters set to $\beta_1=0.9$ and $\beta_2=0.95$. 
The learning rate is set to $1e^{-6}$ for full fine-tuning and $1e^{-6}$ for the others, accompanied by a weight decay of 0.1 and $\epsilon=0.1$ for the smoothing factor.
We set $\alpha=0.1$ and $\lambda=0.001$ to adjust the impact of auxiliary losses.
All experiments are run for 3 epochs. 
Each experiment is repeated three times with different random seeds, and the mean values are reported.
The random seed is shared by all compared methods for a fair comparison.
\subsection{Results}

\subsubsection{Full Data.}
Experimental results in Table \ref{full-data} indicate that \name significantly outperforms both single-dialect and multi-dialect full fine-tuning over nearly 3-5 percent on EX and EXEC in the full data setting. Additionally, \name shows consistent improvement as the model size increases from 1.5B to 14B. Notably, for the 1.5B model, the EX for MySQL and nGQL improves by nearly 10 percent. 
This demonstrates the capability of \name to effectively isolate dialect-specific knowledge and leverage generalized knowledge across multiple dialects for improved performance. Meanwhile, \name is robust enough to achieve stable improvements across various backbones.

\subsubsection{Imbalanced Data.}
We validate \name's transfer capabilities in two data imbalance settings. The first is the MySQL high-resource setting, where we sample the entire MySQL training data and 128 samples from each of the other three dialects.
As shown in Table \ref{mysql-rich}, even under this setting, our method still yields stable better performance over full fine-tuning and LoRA. In particular, on the 14B model size, the average EX of the four dialects is improved by nearly 5 percent on average.

The second is the Cypher high-resource setting, where we sample the entire Cypher training data and 128 samples from each of the other three dialects. As shown in Table \ref{cypher-rich}, \name generally outperforms other methods. Notably, it achieves the highest average EX for the 1.5B, 7B, and 14B parameter models, particularly excelling in Cypher and nGQL due to their high syntactical similarity.
It facilitates the transfer of extensive dialect-common knowledge from Cypher to nGQL. 
However, other methods occasionally perform better in EX, e.g., MySQL and PG. There are two primary reasons why \name does not fully surpass performance on relational database dialects. First, the training data for relational dialects is insufficient, and the sparsity of the dialect expert group structure leads to incomplete convergence for some experts. Second, the differences between relational and non-relational databases are more pronounced, resulting in less transferable dialect-common knowledge across dialects.

\begin{figure*}[ht]
\centering
\includegraphics[width=0.8\linewidth]{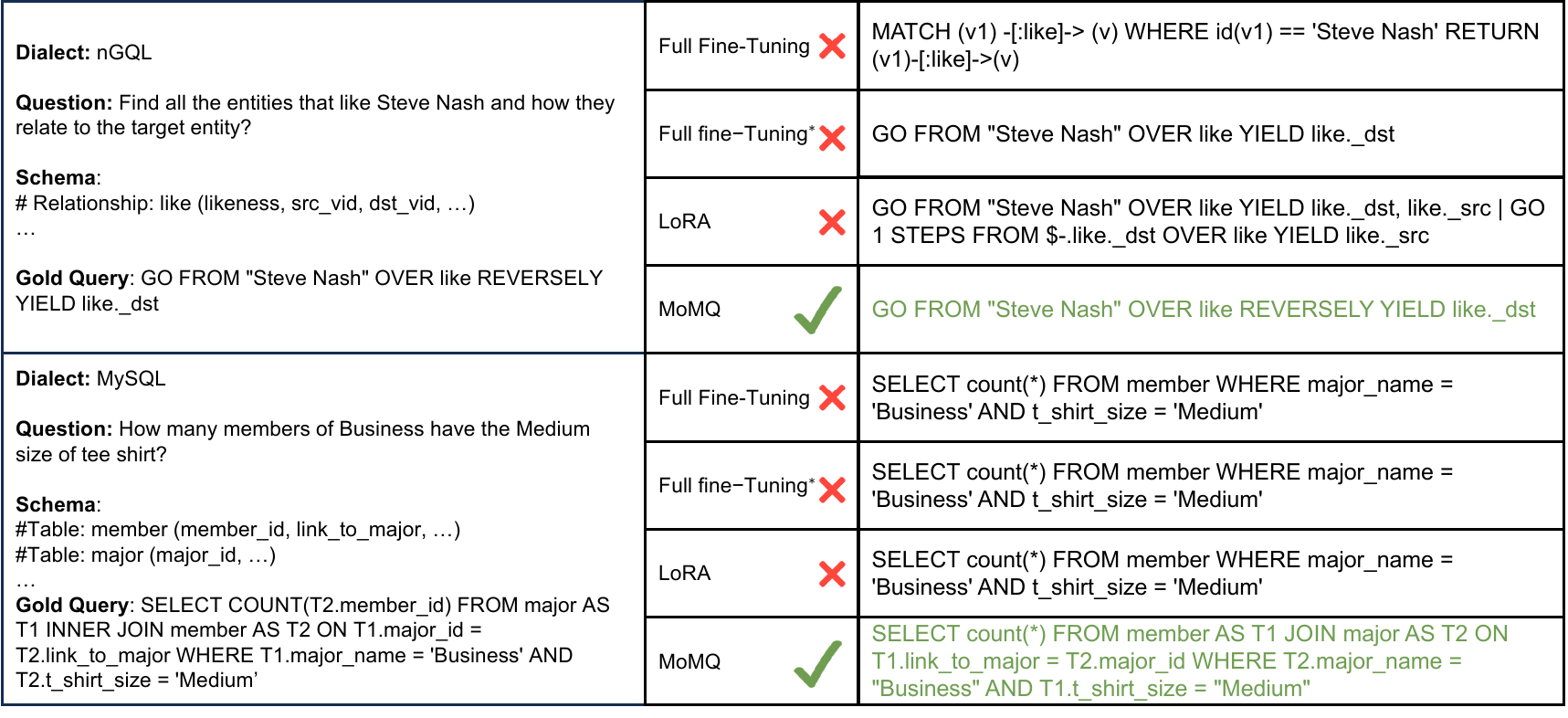}
\caption{Case study of generating nGQL and MySQL queries of different methods in the full data setting.}
\label{case}
\end{figure*}

\subsection{Analysis}
To further validate the effectiveness, robustness, and interpretability of \name, we design a variety of analytical experiments. All experiments are conducted using the Qwen2-7B backbone in the full data setting.
\subsubsection{Ablation Study.}
To evaluate the effectiveness and impact of different components of \name, we conduct ablation studies on Qwen2-7B. 
As shown in Table \ref{ablation}, we remove different components from \name and evaluate the average EX. 
Upon conducting the ablations, notable decreases in performance are observed. 
Specifically, the exclusion of the shared expert group results in a drop in average EX from 49.15\% to 48.57\%. 
Further removal of the dialect router loss leads to an additional decline, with the average EX decreasing to 47.08\%. In this case, tokens are routed with hard dialect labels at the sentence level, resulting in complete dialect isolation.
The most significant reduction occurs when the dialect router is removed, resulting in an average EX of 44.79\%. In this case, tokens are randomly assigned to the dialect expert groups without any supervision.
Finally, eliminating dialect expert groups causes the entire structure to revert to LoRA, further reducing the EX to 43.48\%.
These results underscore the critical contributions of the shared expert group, dialect router loss, dialect router, and dialect expert group to the overall performance of \name.
\begin{table}[h]
    \centering
    \scalebox{1}{
    \begin{tabular}{c|ccccc}
        \Xhline{1pt}
        \textbf{Component} & \multicolumn{5}{c}{\textbf{Ablation}} \\
        \Xhline{1pt}
        % Expert Balance Loss &\checkmark&\times&\times&\times&\times\\
        Shared Expert Group &\checkmark&$\times$&$\times$&$\times$&$\times$\\
        Dialect Router Loss &\checkmark&\checkmark&$\times$&$\times$&$\times$\\
        Dialect Router &\checkmark&\checkmark&\checkmark&$\times$&$\times$\\
        Dialect Expert Group &\checkmark&\checkmark&\checkmark&\checkmark&$\times$\\
        \hline
        \textbf{Avg. EX} & \textbf{49.15} &48.57&47.08&44.79&43.48\\ 
        \Xhline{1pt}
    \end{tabular}
    }
    \caption{Results of ablation studies on Qwen2-7B.}
    \label{ablation}
\end{table}
\begin{table}
    \centering
    \scalebox{1}{
    \begin{tabular}{c|cccccc}
        \Xhline{1pt}
        \textbf{Expert} & \multicolumn{5}{c}{\textbf{Execution Accuracy}} \\
        \cline{2-6}
        \textbf{Dimension}& \textbf{MySQL} & \textbf{PG} & \textbf{Cypher} & \textbf{nGQL} & \textbf{Avg.} \\
        \Xhline{1pt}
        16 & 65.56 & 44.33 & 39.93 & 34.49 & 46.08 \\
        64 & 64.95 & 46.10 & \textbf{42.82} & 36.00 & 47.47 \\
        128 & \textbf{66.30} & 48.12 & 40.97 & \textbf{41.20} & \textbf{49.15} \\
        256 & 65.19 & \textbf{48.23} & 41.20 &39.24 & 48.46 \\
        \Xhline{1pt}
    \end{tabular}
    }
    \caption{Results with different expert dimensions on Qwen2-7B.}
    \label{expert_dim}
\end{table}
\subsubsection{Effect of Expert Dimension.}
We further analyze the impact of the expert dimension on the performance of \name, where the expert dimension refers to the LoRA module's rank. 
As illustrated in Table \ref{expert_dim}, \name demonstrates a consistent increase in average EX as the expert dimension increases from 16 to 128. 
Specifically, with an expert dimension of 128, \name achieves the highest average EX of 49.15\%. It is interesting to note that when the expert dimension is further increased to 256, there is a slight decrease in the average EX. 
This decline suggests a potential issue where a large expert dimension may lead to inadequate training within 3 epochs and consequently impact \name's performance negatively.

\subsubsection{Effect of Expert Number.}
To comprehensively analyze the impact of the number of experts, we evaluate \name with configurations of 8, 16, 32, and 64 experts, respectively. 
As shown in Table \ref{expert_number}, \name demonstrates robust performance across all configurations, particularly excelling with 8 and 32 experts. 
When the number of experts reaches 64, there is also a certain decline in \name's performance. 
This is attributed to a similar reason as when the expert dimension reaches 256, indicating that the experts are not sufficiently trained.
\begin{table}
    \centering
    \scalebox{1}{
    \begin{tabular}{c|ccccc}
        \Xhline{1pt}
        \textbf{Expert} & \multicolumn{5}{c}{\textbf{Execution Accuracy}} \\
        \cline{2-6}
        \textbf{Number}& \textbf{MySQL} & \textbf{PG} & \textbf{Cypher} & \textbf{nGQL} & \textbf{Avg.} \\
        \Xhline{1pt}
        8 & \textbf{66.67} & 46.93 & \textbf{42.25} & 38.77 & 48.65 \\
        16 & 65.68 & 48.11 & 40.74 & 38.19 & 48.18 \\
        32 & 66.30 & \textbf{48.12} & 40.97 & \textbf{41.20} & \textbf{49.15} \\
        64 & 65.68 & 47.40 & 37.62 & 38.89 & 47.40 \\
        \Xhline{1pt}
    \end{tabular}
    }
    \caption{Results with different expert numbers on Qwen2-7B.}
    \label{expert_number}
\end{table}
\subsubsection{Case Study.}
Figure \ref{case} shows the comparison of different methods for generating nGQL and MySQL queries.
From the above results, only \name generates the correct queries.
For nGQL dialect, the multi-dialect full fine-tuning method uses the "MATCH" statement instead of the "GO FROM" statement, which is interfered by the Cypher syntax.
Single-dialect full fine-tuning and LoRA methods are unable to generate the correct query, although it uses the right nGQL syntax.
For MySQL dialect, both full fine-tuning and LoRA do not perform a join operation on the table, indicating a lack of understanding of the question and schema.
These cases further illustrate that \name has a stronger resistance to interference and a better understanding of common information compared to other methods.

\begin{figure}[tb]
\centering
\includegraphics[width=1\linewidth]{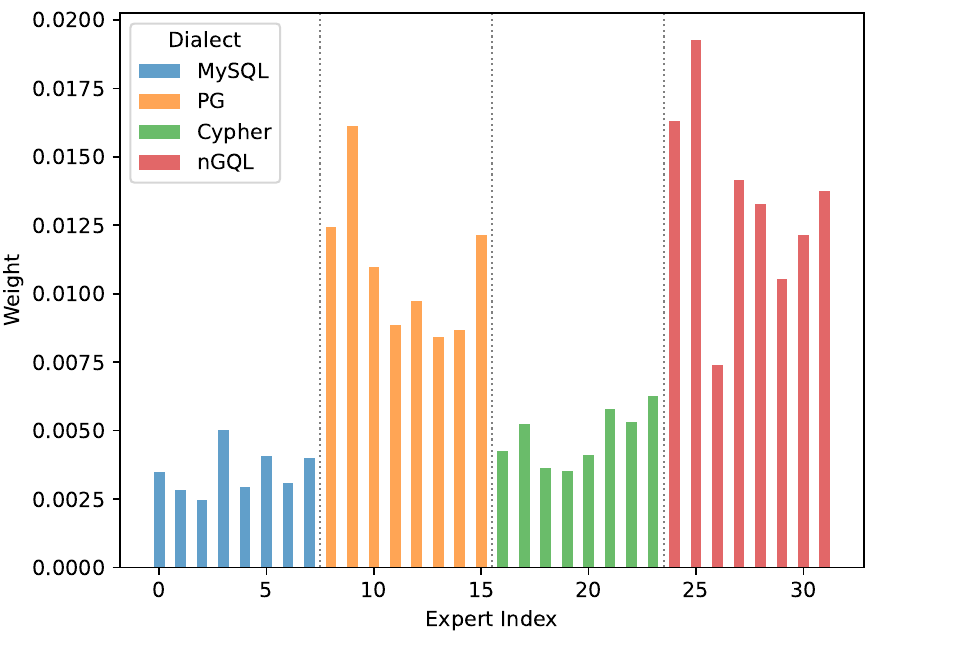}
\caption{Expert weight distribution of generating the nGQL query. A certain number of experts are activated in each expert group and the nGQL expert group plays a dominant role in this generation process.}
\label{weight}
\end{figure}
\subsubsection{Expert Weight Distribution.}

To further analyze the effect of the multi-level routing strategy, we collect the output logits of the dialect router and the expert router from all Transformer layers when generating the nGQL query in the case study. As illustrated in Figure \ref{weight}, under the constraint of the Expert Balance Loss, expert weights within each dialect expert group remain balanced after training.
At the same time, each dialect expert group has activated experts, indicating that the expert groups are not entirely isolated. Tokens from the nGQL dialect have the opportunity to influence other expert groups, suggesting a degree of knowledge transfer. Furthermore, we observe that the weights of the nGQL expert group are significantly greater than those of the other groups, demonstrating that the nGQL experts play a dominant role in this generation process. This observation further validates the effectiveness of the Dialect Router Loss.

%% file: section/conclusion.tex
\section{Conclusion}
In this paper, we propose \name, a novel Mixture-of-Experts-based multi-dialect query generation framework across relational and non-relational databases. 
\name employs specific dialect expert groups for each dialect to isolate dialect-specific knowledge and mitigate generation interference. To deal with multi-dialect data imbalance, we introduce a shared expert group to enhance the transfer of common knowledge from high-resource dialects to low-resource dialects.
% We propose a novel multi-level routing strategy and Dialect Router Loss to further achieve implicit information isolation and fusion. 
We further design a multi-level routing strategy that consists of a dialect router and an expert router to ensure correct routing at the token level. The dialect router enhances knowledge transfer across expert groups with the help of the Dialect Router Loss.
We have constructed a high-quality multi-dialect dataset
covering MySQL, PostgreSQL, Cypher for Neo4j, and nGQL for NebulaGraph.
Both the code and multi-dialect datasets will be openly released to support further research in this area.

\section*{Ethical Considerations}
The deployment of database query generation systems raises several ethical considerations, particularly regarding data privacy, security, and misuse. First and foremost, such systems must ensure that sensitive data is protected from unauthorized access. As these models parse and generate queries from natural language, there is a risk of inadvertently exposing confidential information if proper safeguards are not in place. Additionally, the potential for generating harmful or malicious queries that could compromise database integrity or lead to data breaches must be addressed.

Moreover, considerations surrounding user intent and the context of queries are crucial. Misinterpretations between the user's natural language input and the intended query commands could lead to unintended actions, such as the alteration or deletion of critical data. It is essential to implement robust validation mechanisms to mitigate these risks. Lastly, transparency in the model's decision-making process is vital to ensure accountability and trust. Users should be made aware of how their queries are processed and the limitations of the system. By addressing these ethical considerations, developers and researchers can better align database query generation technologies with responsible use and societal values.

%% file: sample-sigconf.bbl

\begin{thebibliography}{38}

%%% the Web address.

\ifx \showCODEN    \undefined \def \showCODEN     #1{\unskip}     \fi
\ifx \showDOI      \undefined \def \showDOI       #1{#1}\fi
\ifx \showISBNx    \undefined \def \showISBNx     #1{\unskip}     \fi
\ifx \showISBNxiii \undefined \def \showISBNxiii  #1{\unskip}     \fi
\ifx \showISSN     \undefined \def \showISSN      #1{\unskip}     \fi
\ifx \showLCCN     \undefined \def \showLCCN      #1{\unskip}     \fi
\ifx \shownote     \undefined \def \shownote      #1{#1}          \fi
\ifx \showarticletitle \undefined \def \showarticletitle #1{#1}   \fi
\ifx \showURL      \undefined \def \showURL       {\relax}        \fi
% invisible to TeX
\providecommand\bibfield[2]{#2}
\providecommand\bibinfo[2]{#2}
\providecommand\natexlab[1]{#1}
\providecommand\showeprint[2][]{arXiv:#2}

\bibitem[Achiam et~al\mbox{.}(2023)]%
        {achiam2023gpt}
\bibfield{author}{\bibinfo{person}{Josh Achiam}, \bibinfo{person}{Steven Adler}, \bibinfo{person}{Sandhini Agarwal}, \bibinfo{person}{Lama Ahmad}, \bibinfo{person}{Ilge Akkaya}, \bibinfo{person}{Florencia~Leoni Aleman}, \bibinfo{person}{Diogo Almeida}, \bibinfo{person}{Janko Altenschmidt}, \bibinfo{person}{Sam Altman}, \bibinfo{person}{Shyamal Anadkat}, {et~al\mbox{.}}} \bibinfo{year}{2023}\natexlab{}.
\newblock \showarticletitle{Gpt-4 technical report}.
\newblock \bibinfo{journal}{\emph{arXiv preprint arXiv:2303.08774}} (\bibinfo{year}{2023}).
\newblock


\bibitem[Anil et~al\mbox{.}(2023)]%
        {anil2023palm}
\bibfield{author}{\bibinfo{person}{Rohan Anil}, \bibinfo{person}{Andrew~M Dai}, \bibinfo{person}{Orhan Firat}, \bibinfo{person}{Melvin Johnson}, \bibinfo{person}{Dmitry Lepikhin}, \bibinfo{person}{Alexandre Passos}, \bibinfo{person}{Siamak Shakeri}, \bibinfo{person}{Emanuel Taropa}, \bibinfo{person}{Paige Bailey}, \bibinfo{person}{Zhifeng Chen}, {et~al\mbox{.}}} \bibinfo{year}{2023}\natexlab{}.
\newblock \showarticletitle{Palm 2 technical report}.
\newblock \bibinfo{journal}{\emph{arXiv preprint arXiv:2305.10403}} (\bibinfo{year}{2023}).
\newblock


\bibitem[Bai et~al\mbox{.}(2023)]%
        {bai2023qwen}
\bibfield{author}{\bibinfo{person}{Jinze Bai}, \bibinfo{person}{Shuai Bai}, \bibinfo{person}{Yunfei Chu}, \bibinfo{person}{Zeyu Cui}, \bibinfo{person}{Kai Dang}, \bibinfo{person}{Xiaodong Deng}, \bibinfo{person}{Yang Fan}, \bibinfo{person}{Wenbin Ge}, \bibinfo{person}{Yu Han}, \bibinfo{person}{Fei Huang}, {et~al\mbox{.}}} \bibinfo{year}{2023}\natexlab{}.
\newblock \showarticletitle{Qwen technical report}.
\newblock \bibinfo{journal}{\emph{arXiv preprint arXiv:2309.16609}} (\bibinfo{year}{2023}).
\newblock


\bibitem[Crawshaw(2020)]%
        {multitask_better3}
\bibfield{author}{\bibinfo{person}{Michael Crawshaw}.} \bibinfo{year}{2020}\natexlab{}.
\newblock \bibinfo{title}{Multi-Task Learning with Deep Neural Networks: A Survey}.
\newblock
\newblock
\showeprint[arxiv]{2009.09796}~[cs.LG]
\urldef\tempurl%
\url{https://arxiv.org/abs/2009.09796}
\showURL{%
\tempurl}


\bibitem[Dai et~al\mbox{.}(2024)]%
        {dai2024deepseekmoeultimateexpertspecialization}
\bibfield{author}{\bibinfo{person}{Damai Dai}, \bibinfo{person}{Chengqi Deng}, \bibinfo{person}{Chenggang Zhao}, \bibinfo{person}{R.~X. Xu}, \bibinfo{person}{Huazuo Gao}, \bibinfo{person}{Deli Chen}, \bibinfo{person}{Jiashi Li}, \bibinfo{person}{Wangding Zeng}, \bibinfo{person}{Xingkai Yu}, \bibinfo{person}{Y. Wu}, \bibinfo{person}{Zhenda Xie}, \bibinfo{person}{Y.~K. Li}, \bibinfo{person}{Panpan Huang}, \bibinfo{person}{Fuli Luo}, \bibinfo{person}{Chong Ruan}, \bibinfo{person}{Zhifang Sui}, {and} \bibinfo{person}{Wenfeng Liang}.} \bibinfo{year}{2024}\natexlab{}.
\newblock \bibinfo{title}{DeepSeekMoE: Towards Ultimate Expert Specialization in Mixture-of-Experts Language Models}.
\newblock
\newblock
\showeprint[arxiv]{2401.06066}~[cs.CL]
\urldef\tempurl%
\url{https://arxiv.org/abs/2401.06066}
\showURL{%
\tempurl}


\bibitem[Dou et~al\mbox{.}(2023)]%
        {dou2023multispider}
\bibfield{author}{\bibinfo{person}{Longxu Dou}, \bibinfo{person}{Yan Gao}, \bibinfo{person}{Mingyang Pan}, \bibinfo{person}{Dingzirui Wang}, \bibinfo{person}{Wanxiang Che}, \bibinfo{person}{Dechen Zhan}, {and} \bibinfo{person}{Jian-Guang Lou}.} \bibinfo{year}{2023}\natexlab{}.
\newblock \showarticletitle{MultiSpider: towards benchmarking multilingual text-to-SQL semantic parsing}. In \bibinfo{booktitle}{\emph{Proceedings of the AAAI Conference on Artificial Intelligence}}, Vol.~\bibinfo{volume}{37}. \bibinfo{pages}{12745--12753}.
\newblock


\bibitem[Du et~al\mbox{.}(2022)]%
        {pmlr-v162-du22c}
\bibfield{author}{\bibinfo{person}{Nan Du}, \bibinfo{person}{Yanping Huang}, \bibinfo{person}{Andrew~M Dai}, \bibinfo{person}{Simon Tong}, \bibinfo{person}{Dmitry Lepikhin}, \bibinfo{person}{Yuanzhong Xu}, \bibinfo{person}{Maxim Krikun}, \bibinfo{person}{Yanqi Zhou}, \bibinfo{person}{Adams~Wei Yu}, \bibinfo{person}{Orhan Firat}, \bibinfo{person}{Barret Zoph}, \bibinfo{person}{Liam Fedus}, \bibinfo{person}{Maarten~P Bosma}, \bibinfo{person}{Zongwei Zhou}, \bibinfo{person}{Tao Wang}, \bibinfo{person}{Emma Wang}, \bibinfo{person}{Kellie Webster}, \bibinfo{person}{Marie Pellat}, \bibinfo{person}{Kevin Robinson}, \bibinfo{person}{Kathleen Meier-Hellstern}, \bibinfo{person}{Toju Duke}, \bibinfo{person}{Lucas Dixon}, \bibinfo{person}{Kun Zhang}, \bibinfo{person}{Quoc Le}, \bibinfo{person}{Yonghui Wu}, \bibinfo{person}{Zhifeng Chen}, {and} \bibinfo{person}{Claire Cui}.} \bibinfo{year}{2022}\natexlab{}.
\newblock \showarticletitle{{GL}a{M}: Efficient Scaling of Language Models with Mixture-of-Experts}. In \bibinfo{booktitle}{\emph{Proceedings of the 39th International Conference on Machine Learning}} \emph{(\bibinfo{series}{Proceedings of Machine Learning Research}, Vol.~\bibinfo{volume}{162})}, \bibfield{editor}{\bibinfo{person}{Kamalika Chaudhuri}, \bibinfo{person}{Stefanie Jegelka}, \bibinfo{person}{Le~Song}, \bibinfo{person}{Csaba Szepesvari}, \bibinfo{person}{Gang Niu}, {and} \bibinfo{person}{Sivan Sabato}} (Eds.). \bibinfo{publisher}{PMLR}, \bibinfo{pages}{5547--5569}.
\newblock
\urldef\tempurl%
\url{https://proceedings.mlr.press/v162/du22c.html}
\showURL{%
\tempurl}


\bibitem[Fedus et~al\mbox{.}(2022)]%
        {fedus2022switchtransformersscalingtrillion}
\bibfield{author}{\bibinfo{person}{William Fedus}, \bibinfo{person}{Barret Zoph}, {and} \bibinfo{person}{Noam Shazeer}.} \bibinfo{year}{2022}\natexlab{}.
\newblock \bibinfo{title}{Switch Transformers: Scaling to Trillion Parameter Models with Simple and Efficient Sparsity}.
\newblock
\newblock
\showeprint[arxiv]{2101.03961}~[cs.LG]
\urldef\tempurl%
\url{https://arxiv.org/abs/2101.03961}
\showURL{%
\tempurl}


\bibitem[Feng et~al\mbox{.}(2024)]%
        {feng2024mixtureoflorasefficientmultitasktuning}
\bibfield{author}{\bibinfo{person}{Wenfeng Feng}, \bibinfo{person}{Chuzhan Hao}, \bibinfo{person}{Yuewei Zhang}, \bibinfo{person}{Yu Han}, {and} \bibinfo{person}{Hao Wang}.} \bibinfo{year}{2024}\natexlab{}.
\newblock \bibinfo{title}{Mixture-of-LoRAs: An Efficient Multitask Tuning for Large Language Models}.
\newblock
\newblock
\showeprint[arxiv]{2403.03432}~[cs.CL]
\urldef\tempurl%
\url{https://arxiv.org/abs/2403.03432}
\showURL{%
\tempurl}


\bibitem[Fu et~al\mbox{.}(2023)]%
        {fu2023catsql}
\bibfield{author}{\bibinfo{person}{Han Fu}, \bibinfo{person}{Chang Liu}, \bibinfo{person}{Bin Wu}, \bibinfo{person}{Feifei Li}, \bibinfo{person}{Jian Tan}, {and} \bibinfo{person}{Jianling Sun}.} \bibinfo{year}{2023}\natexlab{}.
\newblock \showarticletitle{Catsql: Towards real world natural language to sql applications}.
\newblock \bibinfo{journal}{\emph{Proceedings of the VLDB Endowment}} \bibinfo{volume}{16}, \bibinfo{number}{6} (\bibinfo{year}{2023}), \bibinfo{pages}{1534--1547}.
\newblock


\bibitem[Gao et~al\mbox{.}(2024)]%
        {dail-sql}
\bibfield{author}{\bibinfo{person}{Dawei Gao}, \bibinfo{person}{Haibin Wang}, \bibinfo{person}{Yaliang Li}, \bibinfo{person}{Xiuyu Sun}, \bibinfo{person}{Yichen Qian}, \bibinfo{person}{Bolin Ding}, {and} \bibinfo{person}{Jingren Zhou}.} \bibinfo{year}{2024}\natexlab{}.
\newblock \showarticletitle{Text-to-SQL Empowered by Large Language Models: A Benchmark Evaluation}.
\newblock \bibinfo{journal}{\emph{Proc. VLDB Endow.}} \bibinfo{volume}{17}, \bibinfo{number}{5} (\bibinfo{date}{may} \bibinfo{year}{2024}), \bibinfo{pages}{1132–1145}.
\newblock
\showISSN{2150-8097}
\urldef\tempurl%
\url{https://doi.org/10.14778/3641204.3641221}
\showDOI{\tempurl}


\bibitem[Guo et~al\mbox{.}(2021)]%
        {guo-etal-2021-chase}
\bibfield{author}{\bibinfo{person}{Jiaqi Guo}, \bibinfo{person}{Ziliang Si}, \bibinfo{person}{Yu Wang}, \bibinfo{person}{Qian Liu}, \bibinfo{person}{Ming Fan}, \bibinfo{person}{Jian-Guang Lou}, \bibinfo{person}{Zijiang Yang}, {and} \bibinfo{person}{Ting Liu}.} \bibinfo{year}{2021}\natexlab{}.
\newblock \showarticletitle{Chase: A Large-Scale and Pragmatic {C}hinese Dataset for Cross-Database Context-Dependent Text-to-{SQL}}. In \bibinfo{booktitle}{\emph{Proceedings of the 59th Annual Meeting of the Association for Computational Linguistics and the 11th International Joint Conference on Natural Language Processing (Volume 1: Long Papers)}}, \bibfield{editor}{\bibinfo{person}{Chengqing Zong}, \bibinfo{person}{Fei Xia}, \bibinfo{person}{Wenjie Li}, {and} \bibinfo{person}{Roberto Navigli}} (Eds.). \bibinfo{publisher}{Association for Computational Linguistics}, \bibinfo{address}{Online}, \bibinfo{pages}{2316--2331}.
\newblock
\urldef\tempurl%
\url{https://doi.org/10.18653/v1/2021.acl-long.180}
\showDOI{\tempurl}


\bibitem[Hu et~al\mbox{.}(2021)]%
        {hu2021lora}
\bibfield{author}{\bibinfo{person}{Edward~J Hu}, \bibinfo{person}{Yelong Shen}, \bibinfo{person}{Phillip Wallis}, \bibinfo{person}{Zeyuan Allen-Zhu}, \bibinfo{person}{Yuanzhi Li}, \bibinfo{person}{Shean Wang}, \bibinfo{person}{Lu Wang}, {and} \bibinfo{person}{Weizhu Chen}.} \bibinfo{year}{2021}\natexlab{}.
\newblock \showarticletitle{Lora: Low-rank adaptation of large language models}.
\newblock \bibinfo{journal}{\emph{arXiv preprint arXiv:2106.09685}} (\bibinfo{year}{2021}).
\newblock


\bibitem[Jiang et~al\mbox{.}(2024)]%
        {jiang2024mixtralexperts}
\bibfield{author}{\bibinfo{person}{Albert~Q. Jiang}, \bibinfo{person}{Alexandre Sablayrolles}, \bibinfo{person}{Antoine Roux}, \bibinfo{person}{Arthur Mensch}, \bibinfo{person}{Blanche Savary}, \bibinfo{person}{Chris Bamford}, \bibinfo{person}{Devendra~Singh Chaplot}, \bibinfo{person}{Diego de~las Casas}, \bibinfo{person}{Emma~Bou Hanna}, \bibinfo{person}{Florian Bressand}, \bibinfo{person}{Gianna Lengyel}, \bibinfo{person}{Guillaume Bour}, \bibinfo{person}{Guillaume Lample}, \bibinfo{person}{Lélio~Renard Lavaud}, \bibinfo{person}{Lucile Saulnier}, \bibinfo{person}{Marie-Anne Lachaux}, \bibinfo{person}{Pierre Stock}, \bibinfo{person}{Sandeep Subramanian}, \bibinfo{person}{Sophia Yang}, \bibinfo{person}{Szymon Antoniak}, \bibinfo{person}{Teven~Le Scao}, \bibinfo{person}{Théophile Gervet}, \bibinfo{person}{Thibaut Lavril}, \bibinfo{person}{Thomas Wang}, \bibinfo{person}{Timothée Lacroix}, {and} \bibinfo{person}{William~El Sayed}.} \bibinfo{year}{2024}\natexlab{}.
\newblock \bibinfo{title}{Mixtral of Experts}.
\newblock
\newblock
\showeprint[arxiv]{2401.04088}~[cs.LG]
\urldef\tempurl%
\url{https://arxiv.org/abs/2401.04088}
\showURL{%
\tempurl}


\bibitem[Komatsuzaki et~al\mbox{.}(2022)]%
        {komatsuzakisparse}
\bibfield{author}{\bibinfo{person}{Aran Komatsuzaki}, \bibinfo{person}{Joan Puigcerver}, \bibinfo{person}{James Lee-Thorp}, \bibinfo{person}{Carlos~Riquelme Ruiz}, \bibinfo{person}{Basil Mustafa}, \bibinfo{person}{Joshua Ainslie}, \bibinfo{person}{Yi Tay}, \bibinfo{person}{Mostafa Dehghani}, {and} \bibinfo{person}{Neil Houlsby}.} \bibinfo{year}{2022}\natexlab{}.
\newblock \showarticletitle{Sparse Upcycling: Training Mixture-of-Experts from Dense Checkpoints}. In \bibinfo{booktitle}{\emph{The Eleventh International Conference on Learning Representations}}.
\newblock


\bibitem[Lepikhin et~al\mbox{.}(2020)]%
        {lepikhin2020gshardscalinggiantmodels}
\bibfield{author}{\bibinfo{person}{Dmitry Lepikhin}, \bibinfo{person}{HyoukJoong Lee}, \bibinfo{person}{Yuanzhong Xu}, \bibinfo{person}{Dehao Chen}, \bibinfo{person}{Orhan Firat}, \bibinfo{person}{Yanping Huang}, \bibinfo{person}{Maxim Krikun}, \bibinfo{person}{Noam Shazeer}, {and} \bibinfo{person}{Zhifeng Chen}.} \bibinfo{year}{2020}\natexlab{}.
\newblock \bibinfo{title}{GShard: Scaling Giant Models with Conditional Computation and Automatic Sharding}.
\newblock
\newblock
\showeprint[arxiv]{2006.16668}~[cs.CL]
\urldef\tempurl%
\url{https://arxiv.org/abs/2006.16668}
\showURL{%
\tempurl}


\bibitem[Li et~al\mbox{.}(2024b)]%
        {li2024mixloraenhancinglargelanguage}
\bibfield{author}{\bibinfo{person}{Dengchun Li}, \bibinfo{person}{Yingzi Ma}, \bibinfo{person}{Naizheng Wang}, \bibinfo{person}{Zhengmao Ye}, \bibinfo{person}{Zhiyuan Cheng}, \bibinfo{person}{Yinghao Tang}, \bibinfo{person}{Yan Zhang}, \bibinfo{person}{Lei Duan}, \bibinfo{person}{Jie Zuo}, \bibinfo{person}{Cal Yang}, {and} \bibinfo{person}{Mingjie Tang}.} \bibinfo{year}{2024}\natexlab{b}.
\newblock \bibinfo{title}{MixLoRA: Enhancing Large Language Models Fine-Tuning with LoRA-based Mixture of Experts}.
\newblock
\newblock
\showeprint[arxiv]{2404.15159}~[cs.CL]
\urldef\tempurl%
\url{https://arxiv.org/abs/2404.15159}
\showURL{%
\tempurl}


\bibitem[Li et~al\mbox{.}(2023c)]%
        {resdsql}
\bibfield{author}{\bibinfo{person}{Haoyang Li}, \bibinfo{person}{Jing Zhang}, \bibinfo{person}{Cuiping Li}, {and} \bibinfo{person}{Hong Chen}.} \bibinfo{year}{2023}\natexlab{c}.
\newblock \showarticletitle{RESDSQL: decoupling schema linking and skeleton parsing for text-to-SQL}. In \bibinfo{booktitle}{\emph{Proceedings of the Thirty-Seventh AAAI Conference on Artificial Intelligence and Thirty-Fifth Conference on Innovative Applications of Artificial Intelligence and Thirteenth Symposium on Educational Advances in Artificial Intelligence}} \emph{(\bibinfo{series}{AAAI'23/IAAI'23/EAAI'23})}. \bibinfo{publisher}{AAAI Press}, Article \bibinfo{articleno}{1466}, \bibinfo{numpages}{9}~pages.
\newblock
\showISBNx{978-1-57735-880-0}
\urldef\tempurl%
\url{https://doi.org/10.1609/aaai.v37i11.26535}
\showDOI{\tempurl}


\bibitem[Li et~al\mbox{.}(2024d)]%
        {10.1145/3654930}
\bibfield{author}{\bibinfo{person}{Haoyang Li}, \bibinfo{person}{Jing Zhang}, \bibinfo{person}{Hanbing Liu}, \bibinfo{person}{Ju Fan}, \bibinfo{person}{Xiaokang Zhang}, \bibinfo{person}{Jun Zhu}, \bibinfo{person}{Renjie Wei}, \bibinfo{person}{Hongyan Pan}, \bibinfo{person}{Cuiping Li}, {and} \bibinfo{person}{Hong Chen}.} \bibinfo{year}{2024}\natexlab{d}.
\newblock \showarticletitle{CodeS: Towards Building Open-source Language Models for Text-to-SQL}.
\newblock \bibinfo{journal}{\emph{Proc. ACM Manag. Data}} \bibinfo{volume}{2}, \bibinfo{number}{3}, Article \bibinfo{articleno}{127} (\bibinfo{date}{may} \bibinfo{year}{2024}), \bibinfo{numpages}{28}~pages.
\newblock
\urldef\tempurl%
\url{https://doi.org/10.1145/3654930}
\showDOI{\tempurl}


\bibitem[Li et~al\mbox{.}(2023b)]%
        {10.1609/aaai.v37i11.26536}
\bibfield{author}{\bibinfo{person}{Jinyang Li}, \bibinfo{person}{Binyuan Hui}, \bibinfo{person}{Reynold Cheng}, \bibinfo{person}{Bowen Qin}, \bibinfo{person}{Chenhao Ma}, \bibinfo{person}{Nan Huo}, \bibinfo{person}{Fei Huang}, \bibinfo{person}{Wenyu Du}, \bibinfo{person}{Luo Si}, {and} \bibinfo{person}{Yongbin Li}.} \bibinfo{year}{2023}\natexlab{b}.
\newblock \showarticletitle{Graphix-T5: mixing pre-trained transformers with graph-aware layers for text-to-SQL parsing}. In \bibinfo{booktitle}{\emph{Proceedings of the Thirty-Seventh AAAI Conference on Artificial Intelligence and Thirty-Fifth Conference on Innovative Applications of Artificial Intelligence and Thirteenth Symposium on Educational Advances in Artificial Intelligence}} \emph{(\bibinfo{series}{AAAI'23/IAAI'23/EAAI'23})}. \bibinfo{publisher}{AAAI Press}, Article \bibinfo{articleno}{1467}, \bibinfo{numpages}{9}~pages.
\newblock
\showISBNx{978-1-57735-880-0}
\urldef\tempurl%
\url{https://doi.org/10.1609/aaai.v37i11.26536}
\showDOI{\tempurl}


\bibitem[Li et~al\mbox{.}(2024a)]%
        {li2024can}
\bibfield{author}{\bibinfo{person}{Jinyang Li}, \bibinfo{person}{Binyuan Hui}, \bibinfo{person}{Ge Qu}, \bibinfo{person}{Jiaxi Yang}, \bibinfo{person}{Binhua Li}, \bibinfo{person}{Bowen Li}, \bibinfo{person}{Bailin Wang}, \bibinfo{person}{Bowen Qin}, \bibinfo{person}{Ruiying Geng}, \bibinfo{person}{Nan Huo}, {et~al\mbox{.}}} \bibinfo{year}{2024}\natexlab{a}.
\newblock \showarticletitle{Can llm already serve as a database interface? a big bench for large-scale database grounded text-to-sqls}.
\newblock \bibinfo{journal}{\emph{Advances in Neural Information Processing Systems}}  \bibinfo{volume}{36} (\bibinfo{year}{2024}).
\newblock


\bibitem[Li et~al\mbox{.}(2023a)]%
        {li2023starcodersourceyou}
\bibfield{author}{\bibinfo{person}{Raymond Li}, \bibinfo{person}{Loubna~Ben Allal}, \bibinfo{person}{Yangtian Zi}, \bibinfo{person}{Niklas Muennighoff}, \bibinfo{person}{Denis Kocetkov}, \bibinfo{person}{Chenghao Mou}, \bibinfo{person}{Marc Marone}, \bibinfo{person}{Christopher Akiki}, \bibinfo{person}{Jia Li}, \bibinfo{person}{Jenny Chim}, \bibinfo{person}{Qian Liu}, \bibinfo{person}{Evgenii Zheltonozhskii}, \bibinfo{person}{Terry~Yue Zhuo}, \bibinfo{person}{Thomas Wang}, \bibinfo{person}{Olivier Dehaene}, \bibinfo{person}{Mishig Davaadorj}, \bibinfo{person}{Joel Lamy-Poirier}, \bibinfo{person}{João Monteiro}, \bibinfo{person}{Oleh Shliazhko}, \bibinfo{person}{Nicolas Gontier}, \bibinfo{person}{Nicholas Meade}, \bibinfo{person}{Armel Zebaze}, \bibinfo{person}{Ming-Ho Yee}, \bibinfo{person}{Logesh~Kumar Umapathi}, \bibinfo{person}{Jian Zhu}, \bibinfo{person}{Benjamin Lipkin}, \bibinfo{person}{Muhtasham Oblokulov}, \bibinfo{person}{Zhiruo Wang}, \bibinfo{person}{Rudra Murthy}, \bibinfo{person}{Jason Stillerman}, \bibinfo{person}{Siva~Sankalp Patel}, \bibinfo{person}{Dmitry Abulkhanov}, \bibinfo{person}{Marco Zocca}, \bibinfo{person}{Manan Dey}, \bibinfo{person}{Zhihan Zhang}, \bibinfo{person}{Nour Fahmy}, \bibinfo{person}{Urvashi Bhattacharyya}, \bibinfo{person}{Wenhao Yu}, \bibinfo{person}{Swayam Singh}, \bibinfo{person}{Sasha Luccioni}, \bibinfo{person}{Paulo Villegas}, \bibinfo{person}{Maxim Kunakov}, \bibinfo{person}{Fedor Zhdanov}, \bibinfo{person}{Manuel Romero}, \bibinfo{person}{Tony Lee}, \bibinfo{person}{Nadav Timor}, \bibinfo{person}{Jennifer Ding}, \bibinfo{person}{Claire Schlesinger}, \bibinfo{person}{Hailey Schoelkopf}, \bibinfo{person}{Jan Ebert}, \bibinfo{person}{Tri Dao}, \bibinfo{person}{Mayank Mishra}, \bibinfo{person}{Alex Gu}, \bibinfo{person}{Jennifer Robinson}, \bibinfo{person}{Carolyn~Jane Anderson}, \bibinfo{person}{Brendan Dolan-Gavitt}, \bibinfo{person}{Danish Contractor}, \bibinfo{person}{Siva Reddy}, \bibinfo{person}{Daniel Fried}, \bibinfo{person}{Dzmitry Bahdanau}, \bibinfo{person}{Yacine Jernite}, \bibinfo{person}{Carlos~Muñoz Ferrandis}, \bibinfo{person}{Sean Hughes}, \bibinfo{person}{Thomas Wolf}, \bibinfo{person}{Arjun Guha}, \bibinfo{person}{Leandro von Werra}, {and} \bibinfo{person}{Harm de Vries}.} \bibinfo{year}{2023}\natexlab{a}.
\newblock \bibinfo{title}{StarCoder: may the source be with you!}
\newblock
\newblock
\showeprint[arxiv]{2305.06161}~[cs.CL]
\urldef\tempurl%
\url{https://arxiv.org/abs/2305.06161}
\showURL{%
\tempurl}


\bibitem[Li et~al\mbox{.}(2024c)]%
        {li2024petsqlpromptenhancedtworoundrefinement}
\bibfield{author}{\bibinfo{person}{Zhishuai Li}, \bibinfo{person}{Xiang Wang}, \bibinfo{person}{Jingjing Zhao}, \bibinfo{person}{Sun Yang}, \bibinfo{person}{Guoqing Du}, \bibinfo{person}{Xiaoru Hu}, \bibinfo{person}{Bin Zhang}, \bibinfo{person}{Yuxiao Ye}, \bibinfo{person}{Ziyue Li}, \bibinfo{person}{Rui Zhao}, {and} \bibinfo{person}{Hangyu Mao}.} \bibinfo{year}{2024}\natexlab{c}.
\newblock \bibinfo{title}{PET-SQL: A Prompt-Enhanced Two-Round Refinement of Text-to-SQL with Cross-consistency}.
\newblock
\newblock
\showeprint[arxiv]{2403.09732}~[cs.CL]
\urldef\tempurl%
\url{https://arxiv.org/abs/2403.09732}
\showURL{%
\tempurl}


\bibitem[Lin et~al\mbox{.}(2024)]%
        {lin2024moe}
\bibfield{author}{\bibinfo{person}{Bin Lin}, \bibinfo{person}{Zhenyu Tang}, \bibinfo{person}{Yang Ye}, \bibinfo{person}{Jiaxi Cui}, \bibinfo{person}{Bin Zhu}, \bibinfo{person}{Peng Jin}, \bibinfo{person}{Junwu Zhang}, \bibinfo{person}{Munan Ning}, {and} \bibinfo{person}{Li Yuan}.} \bibinfo{year}{2024}\natexlab{}.
\newblock \showarticletitle{MoE-LLaVA: Mixture of Experts for Large Vision-Language Models}.
\newblock \bibinfo{journal}{\emph{arXiv preprint arXiv:2401.15947}} (\bibinfo{year}{2024}).
\newblock


\bibitem[Neo4j(2012)]%
        {neo4j}
\bibfield{author}{\bibinfo{person}{Neo4j}.} \bibinfo{year}{2012}\natexlab{}.
\newblock \bibinfo{title}{Neo4j - The World’s Leading Graph Database}.
\newblock
\newblock
\urldef\tempurl%
\url{http://neo4j.org/}
\showURL{%
\tempurl}


\bibitem[Pourreza and Rafiei(2023)]%
        {pourreza2023dinsql}
\bibfield{author}{\bibinfo{person}{Mohammadreza Pourreza} {and} \bibinfo{person}{Davood Rafiei}.} \bibinfo{year}{2023}\natexlab{}.
\newblock \showarticletitle{{DIN}-{SQL}: Decomposed In-Context Learning of Text-to-{SQL} with Self-Correction}. In \bibinfo{booktitle}{\emph{Thirty-seventh Conference on Neural Information Processing Systems}}.
\newblock
\urldef\tempurl%
\url{https://openreview.net/forum?id=p53QDxSIc5}
\showURL{%
\tempurl}


\bibitem[Rozière et~al\mbox{.}(2024)]%
        {rozière2024codellamaopenfoundation}
\bibfield{author}{\bibinfo{person}{Baptiste Rozière}, \bibinfo{person}{Jonas Gehring}, \bibinfo{person}{Fabian Gloeckle}, \bibinfo{person}{Sten Sootla}, \bibinfo{person}{Itai Gat}, \bibinfo{person}{Xiaoqing~Ellen Tan}, \bibinfo{person}{Yossi Adi}, \bibinfo{person}{Jingyu Liu}, \bibinfo{person}{Romain Sauvestre}, \bibinfo{person}{Tal Remez}, \bibinfo{person}{Jérémy Rapin}, \bibinfo{person}{Artyom Kozhevnikov}, \bibinfo{person}{Ivan Evtimov}, \bibinfo{person}{Joanna Bitton}, \bibinfo{person}{Manish Bhatt}, \bibinfo{person}{Cristian~Canton Ferrer}, \bibinfo{person}{Aaron Grattafiori}, \bibinfo{person}{Wenhan Xiong}, \bibinfo{person}{Alexandre Défossez}, \bibinfo{person}{Jade Copet}, \bibinfo{person}{Faisal Azhar}, \bibinfo{person}{Hugo Touvron}, \bibinfo{person}{Louis Martin}, \bibinfo{person}{Nicolas Usunier}, \bibinfo{person}{Thomas Scialom}, {and} \bibinfo{person}{Gabriel Synnaeve}.} \bibinfo{year}{2024}\natexlab{}.
\newblock \bibinfo{title}{Code Llama: Open Foundation Models for Code}.
\newblock
\newblock
\showeprint[arxiv]{2308.12950}~[cs.CL]
\urldef\tempurl%
\url{https://arxiv.org/abs/2308.12950}
\showURL{%
\tempurl}


\bibitem[Shazeer et~al\mbox{.}(2017)]%
        {Sparsely-Gated}
\bibfield{author}{\bibinfo{person}{Noam Shazeer}, \bibinfo{person}{*Azalia Mirhoseini}, \bibinfo{person}{*Krzysztof Maziarz}, \bibinfo{person}{Andy Davis}, \bibinfo{person}{Quoc Le}, \bibinfo{person}{Geoffrey Hinton}, {and} \bibinfo{person}{Jeff Dean}.} \bibinfo{year}{2017}\natexlab{}.
\newblock \showarticletitle{Outrageously Large Neural Networks: The Sparsely-Gated Mixture-of-Experts Layer}. In \bibinfo{booktitle}{\emph{International Conference on Learning Representations}}.
\newblock
\urldef\tempurl%
\url{https://openreview.net/forum?id=B1ckMDqlg}
\showURL{%
\tempurl}


\bibitem[Standley et~al\mbox{.}(2020)]%
        {multitask_better2}
\bibfield{author}{\bibinfo{person}{Trevor Standley}, \bibinfo{person}{Amir Zamir}, \bibinfo{person}{Dawn Chen}, \bibinfo{person}{Leonidas Guibas}, \bibinfo{person}{Jitendra Malik}, {and} \bibinfo{person}{Silvio Savarese}.} \bibinfo{year}{2020}\natexlab{}.
\newblock \showarticletitle{Which Tasks Should Be Learned Together in Multi-task Learning?}. In \bibinfo{booktitle}{\emph{Proceedings of the 37th International Conference on Machine Learning}} \emph{(\bibinfo{series}{Proceedings of Machine Learning Research}, Vol.~\bibinfo{volume}{119})}, \bibfield{editor}{\bibinfo{person}{Hal~Daumé III} {and} \bibinfo{person}{Aarti Singh}} (Eds.). \bibinfo{publisher}{PMLR}, \bibinfo{pages}{9120--9132}.
\newblock
\urldef\tempurl%
\url{https://proceedings.mlr.press/v119/standley20a.html}
\showURL{%
\tempurl}


\bibitem[Touvron et~al\mbox{.}(2023)]%
        {touvron2023llamaopenefficientfoundation}
\bibfield{author}{\bibinfo{person}{Hugo Touvron}, \bibinfo{person}{Thibaut Lavril}, \bibinfo{person}{Gautier Izacard}, \bibinfo{person}{Xavier Martinet}, \bibinfo{person}{Marie-Anne Lachaux}, \bibinfo{person}{Timothée Lacroix}, \bibinfo{person}{Baptiste Rozière}, \bibinfo{person}{Naman Goyal}, \bibinfo{person}{Eric Hambro}, \bibinfo{person}{Faisal Azhar}, \bibinfo{person}{Aurelien Rodriguez}, \bibinfo{person}{Armand Joulin}, \bibinfo{person}{Edouard Grave}, {and} \bibinfo{person}{Guillaume Lample}.} \bibinfo{year}{2023}\natexlab{}.
\newblock \bibinfo{title}{LLaMA: Open and Efficient Foundation Language Models}.
\newblock
\newblock
\showeprint[arxiv]{2302.13971}~[cs.CL]
\urldef\tempurl%
\url{https://arxiv.org/abs/2302.13971}
\showURL{%
\tempurl}


\bibitem[Wu et~al\mbox{.}(2022)]%
        {wu2022nebulagraphopensource}
\bibfield{author}{\bibinfo{person}{Min Wu}, \bibinfo{person}{Xinglu Yi}, \bibinfo{person}{Hui Yu}, \bibinfo{person}{Yu Liu}, {and} \bibinfo{person}{Yujue Wang}.} \bibinfo{year}{2022}\natexlab{}.
\newblock \bibinfo{title}{Nebula Graph: An open source distributed graph database}.
\newblock
\newblock
\showeprint[arxiv]{2206.07278}~[cs.DB]
\urldef\tempurl%
\url{https://arxiv.org/abs/2206.07278}
\showURL{%
\tempurl}


\bibitem[Wu et~al\mbox{.}(2024)]%
        {wu2024mixtureloraexperts}
\bibfield{author}{\bibinfo{person}{Xun Wu}, \bibinfo{person}{Shaohan Huang}, {and} \bibinfo{person}{Furu Wei}.} \bibinfo{year}{2024}\natexlab{}.
\newblock \bibinfo{title}{Mixture of LoRA Experts}.
\newblock
\newblock
\showeprint[arxiv]{2404.13628}~[cs.CL]
\urldef\tempurl%
\url{https://arxiv.org/abs/2404.13628}
\showURL{%
\tempurl}


\bibitem[Xue et~al\mbox{.}(2024)]%
        {xue2024openmoeearlyeffortopen}
\bibfield{author}{\bibinfo{person}{Fuzhao Xue}, \bibinfo{person}{Zian Zheng}, \bibinfo{person}{Yao Fu}, \bibinfo{person}{Jinjie Ni}, \bibinfo{person}{Zangwei Zheng}, \bibinfo{person}{Wangchunshu Zhou}, {and} \bibinfo{person}{Yang You}.} \bibinfo{year}{2024}\natexlab{}.
\newblock \bibinfo{title}{OpenMoE: An Early Effort on Open Mixture-of-Experts Language Models}.
\newblock
\newblock
\showeprint[arxiv]{2402.01739}~[cs.CL]
\urldef\tempurl%
\url{https://arxiv.org/abs/2402.01739}
\showURL{%
\tempurl}


\bibitem[Yu et~al\mbox{.}(2018)]%
        {yu2018spider}
\bibfield{author}{\bibinfo{person}{Tao Yu}, \bibinfo{person}{Rui Zhang}, \bibinfo{person}{Kai Yang}, \bibinfo{person}{Michihiro Yasunaga}, \bibinfo{person}{Dongxu Wang}, \bibinfo{person}{Zifan Li}, \bibinfo{person}{James Ma}, \bibinfo{person}{Irene Li}, \bibinfo{person}{Qingning Yao}, \bibinfo{person}{Shanelle Roman}, {et~al\mbox{.}}} \bibinfo{year}{2018}\natexlab{}.
\newblock \showarticletitle{Spider: A Large-Scale Human-Labeled Dataset for Complex and Cross-Domain Semantic Parsing and Text-to-SQL Task}. In \bibinfo{booktitle}{\emph{Proceedings of the 2018 Conference on Empirical Methods in Natural Language Processing}}. \bibinfo{pages}{3911--3921}.
\newblock


\bibitem[Zhang and Yang(2022)]%
        {multitask_better}
\bibfield{author}{\bibinfo{person}{Yu Zhang} {and} \bibinfo{person}{Qiang Yang}.} \bibinfo{year}{2022}\natexlab{}.
\newblock \showarticletitle{A Survey on Multi-Task Learning}.
\newblock \bibinfo{journal}{\emph{IEEE Transactions on Knowledge and Data Engineering}} \bibinfo{volume}{34}, \bibinfo{number}{12} (\bibinfo{year}{2022}), \bibinfo{pages}{5586--5609}.
\newblock
\urldef\tempurl%
\url{https://doi.org/10.1109/TKDE.2021.3070203}
\showDOI{\tempurl}


\bibitem[Zhao et~al\mbox{.}(2024)]%
        {zhao2024sparse}
\bibfield{author}{\bibinfo{person}{Xinyu Zhao}, \bibinfo{person}{Xuxi Chen}, \bibinfo{person}{Yu Cheng}, {and} \bibinfo{person}{Tianlong Chen}.} \bibinfo{year}{2024}\natexlab{}.
\newblock \showarticletitle{Sparse MoE with Language Guided Routing for Multilingual Machine Translation}. In \bibinfo{booktitle}{\emph{The Twelfth International Conference on Learning Representations}}.
\newblock
\urldef\tempurl%
\url{https://openreview.net/forum?id=ySS7hH1smL}
\showURL{%
\tempurl}


\bibitem[Zhou et~al\mbox{.}(2024)]%
        {zhou2024r3nl2gqlmodelcoordinationknowledge}
\bibfield{author}{\bibinfo{person}{Yuhang Zhou}, \bibinfo{person}{Yu He}, \bibinfo{person}{Siyu Tian}, \bibinfo{person}{Yuchen Ni}, \bibinfo{person}{Zhangyue Yin}, \bibinfo{person}{Xiang Liu}, \bibinfo{person}{Chuanjun Ji}, \bibinfo{person}{Sen Liu}, \bibinfo{person}{Xipeng Qiu}, \bibinfo{person}{Guangnan Ye}, {and} \bibinfo{person}{Hongfeng Chai}.} \bibinfo{year}{2024}\natexlab{}.
\newblock \bibinfo{title}{$R^3$-NL2GQL: A Model Coordination and Knowledge Graph Alignment Approach for NL2GQL}.
\newblock
\newblock
\showeprint[arxiv]{2311.01862}~[cs.CL]
\urldef\tempurl%
\url{https://arxiv.org/abs/2311.01862}
\showURL{%
\tempurl}


\bibitem[Zoph(2022)]%
        {9835248}
\bibfield{author}{\bibinfo{person}{Barret Zoph}.} \bibinfo{year}{2022}\natexlab{}.
\newblock \showarticletitle{Designing Effective Sparse Expert Models}. In \bibinfo{booktitle}{\emph{2022 IEEE International Parallel and Distributed Processing Symposium Workshops (IPDPSW)}}. \bibinfo{pages}{1044--1044}.
\newblock
\urldef\tempurl%
\url{https://doi.org/10.1109/IPDPSW55747.2022.00171}
\showDOI{\tempurl}


\end{thebibliography}
